\documentclass[journal,12pt,onecolumn,draftclsnofoot,peereviewca,twoside]{IEEEtran}

\usepackage{amsmath,amsfonts}
\usepackage{algorithmic}
\usepackage{algorithm}
\usepackage{array}
\usepackage{textcomp}
\usepackage{url}
\usepackage{verbatim}
\usepackage{graphicx}
\usepackage{cite}
\usepackage{float}
\usepackage{bm}
\usepackage{color}
\usepackage{multirow}
\usepackage{multicol}
\usepackage{subcaption}
\usepackage{hyperref}
\usepackage{cite}
\usepackage{graphicx}
\usepackage{subcaption}
\usepackage{stackengine}
\usepackage{tabularx}
\usepackage{dblfloatfix}
\usepackage{tabstackengine}
\usepackage{comment}

\IEEEoverridecommandlockouts                              

\usepackage[dvipsnames]{xcolor}

\DeclareMathOperator*{\argmin}{arg\,min}
\DeclareMathOperator*{\subto}{subject~to~}
\DeclareMathOperator*{\diag}{diag}

\graphicspath{ {./figures/} }

\title{\LARGE \bf
ASV Station Keeping under Wind Disturbances using Neural Network Simulation Error Minimization Model Predictive Control  
 }

\author{Jalil Chavez-Galaviz$^{1}$, Jianwen Li$^{1}$, Ajinkya Chaudhary$^{1}$, and Nina Mahmoudian$^{2}$
\thanks{* This work is partially supported by ONR N00014-20-1-2085.}
\thanks{$^{1}$ J. Chavez-Galaviz, J. Li and A. Chaudhary are Graduate Research Assistants with the School of Mechanical Engineering, Purdue University, West Lafayette, IN, USA. J. Chavez-Galaviz and J. Li contributed equally to this work.
       }
\thanks{$^{2}$ N. Mahmoudian is an associate professor and B.F.S. Schafer Scholar with the School of Mechanical Engineering, Purdue University, West Lafayette, IN, USA.
        {\tt\small ninam@purdue.edu}.}%
}

\begin{document}

\maketitle
\thispagestyle{empty}
\pagestyle{empty}

\abstract[Abstract]{Station keeping is an essential maneuver for Autonomous Surface Vehicles (ASVs), mainly when used in confined spaces, to carry out surveys that require the ASV to keep its position or in collaboration with other vehicles where the relative position has an impact over the mission. However, this maneuver can become challenging for classic feedback controllers due to the need for an accurate model of the ASV dynamics and the environmental disturbances. This work proposes a Model Predictive Controller using Neural Network Simulation Error Minimization (NNSEM-MPC) to accurately predict the dynamics of the ASV under wind disturbances. The performance of the proposed scheme under wind disturbances is tested and compared against other controllers in simulation, using the Robotics Operating System (ROS) and the multipurpose simulation environment Gazebo. A set of six tests were conducted by combining two wind speeds (3 m/s and 6 m/s) and three wind directions (0$^\circ$, 90$^\circ$, and 180$^\circ$). The simulation results clearly show the advantage of the NNSEM-MPC over the following methods: backstepping controller, sliding mode controller, simplified dynamics MPC (SD-MPC), neural ordinary differential equation MPC (NODE-MPC), and knowledge-based NODE MPC (KNODE-MPC). The proposed NNSEM-MPC approach performs better than the rest in 4 out of the 6 test conditions, and it is the second best in the 2 remaining test cases, reducing the mean position and heading error by at least 31\% and 46\% respectively across all the test cases. In terms of execution speed, the proposed NNSEM-MPC is at least 36\% faster than the rest of the MPC controllers. The field experiments on two different ASV platforms showed that ASVs can effectively keep the station utilizing the proposed method, with a position error as low as $1.68$ m and a heading error as low as $6.14^{\circ}$ within time windows of at least $150$s.}


\maketitle

\renewcommand\thefootnote{}

\renewcommand\thefootnote{\fnsymbol{footnote}}
\setcounter{footnote}{1}

\section{INTRODUCTION}
Autonomous surface vehicles (ASVs) have commercial and military applications ranging from patrolling, monitoring, surveying, surveillance, or mapping \cite{qu2022nonlinear}, \cite{sarda2016station}. They can also be used to reduce human intervention during autonomous undersea explorations. They can carry launch and recovery mechanisms for Autonomous Underwater Vehicles (AUVs), charging stations for replenishment, and external sensors such as communication modules for efficient communication or accurate localization to enable efficient long-term missions that can significantly improve how we explore the marine and undersea environment \cite{sarda2018launch,page2021underwater},\cite{li2020collaborative,bresciani2021cooperative,busquets2013auv}.

\begin{figure}[t]
     \centering
     \begin{subfigure}[b]{0.45\textwidth}
         \centering
         \includegraphics[width=\textwidth]{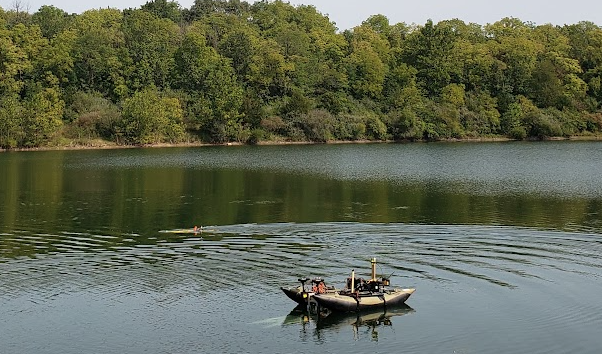}
     \end{subfigure}
     \hspace{-5pt}
     \begin{subfigure}[b]{0.457\textwidth}
         \centering
         \includegraphics[width=\textwidth]{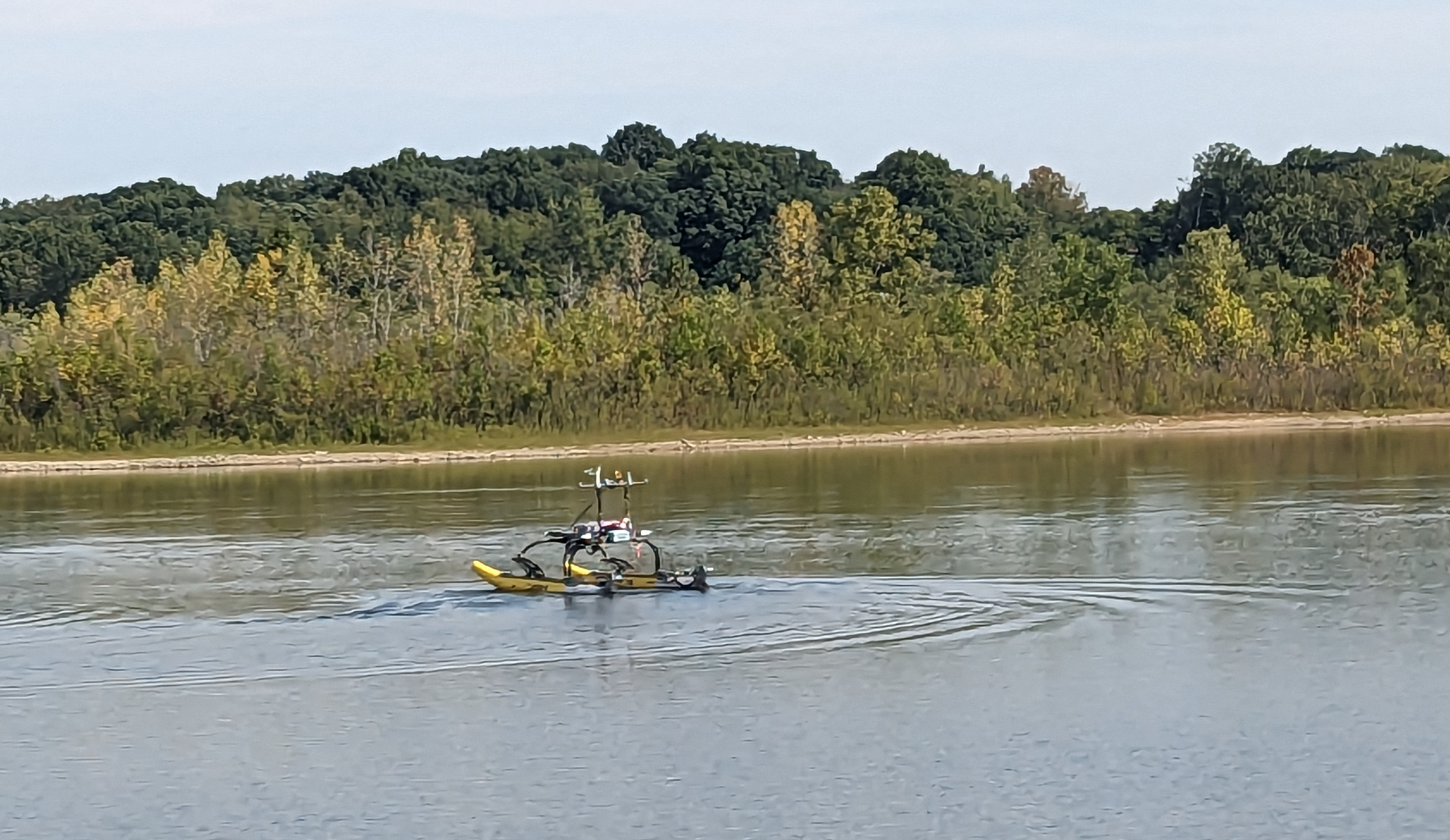}
     \end{subfigure}
        \caption{On the left-hand side the Boat for Robot Engineering and Applied Machine Learning (BREAM) towing a docking station used for the recovery and recharge of an AUV, on the right-hand side, the Wave Adaptive Modular Platform (WAM-V 16). Both platforms were used to implement the proposed MPC approach.}
        \label{fig:asv_operation}
\end{figure}

To encompass the full spectrum of potential applications, Autonomous Surface Vehicles (ASVs) must execute a diverse set of maneuvers, including path tracking, obstacle avoidance, and station keeping \cite{jiang2022data}. Path tracking and obstacle avoidance enable the vehicle to navigate along collision-free trajectories within predefined paths, while station keeping allows the ASV to maintain a specific position and heading for some predefined time. This maneuver significantly enhances the autonomy of ASVs, expanding their utility across diverse scenarios such as biological surveys, maritime rescue missions \cite{qu2022nonlinear, sarda2016station}, Launch and Recovery (LAR) operations \cite{sarda2018launch}, and docking in moderate water conditions. An illustration of a mobile docking station setup, exemplifying the recharging of an Iver3 Autonomous Underwater Vehicle (AUV), is depicted in Fig.~\ref{fig:asv_operation}. Application of station keeping has been effectively demonstrated in \cite{qu2022nonlinear, pereira2008experimental}, in which the maneuver dynamically positions ASVs in specific locations without the need for anchoring.

Classical feedback controllers are common for station keeping of ASVs and have been extensively studied. The existing work can be grouped into those applying robust feedback control to attenuate slow environmental changes \cite{kitamura1997estimation}, and those applying feedforward controllers to counteract rapid changes caused by wind \cite{schlipf2012comparison}. Other work has been focused on applying robust nonlinear control theory such as backstepping or sliding mode controllers for station keeping in combination with feedforward controllers to allow a faster response \cite{sarda2016station,qu2022nonlinear}.

Optimal control techniques such as Model Predictive Control (MPC) have been used to control nonlinear systems. MPC requires an accurate model of the system to achieve good results \cite{seel2021neural}. Since ASVs are nonlinear systems that often operate in challenging environments, a mathematical model is not enough to model all the system dynamics under uncertainties. To overcome this problem, data-driven models such as neural networks, neural ordinary differential equations (NODE), knowledge-based neural ordinary differential equations (KNODE) \cite{xu2022physics,chee2022knode,forgione2021continuous, NEURIPS2018_69386f6b}, or regression models \cite{kaiser2018sparse,seel2021neural}, have become popular to do system identification, which in combination with MPC have shown impressive results mostly in simulation \cite{jiahao2023online,chee2022knode,saint1991neural,kaiser2018sparse,chua2018deep,salzmann2022neural,williams2017information}.

Several works have focused on solving the system identification problem for ASVs using different approaches. The most common one is the white box approach, in which the dynamics of the vehicle are expressed as equations in terms of the forces and moments associated with the ASV motion. This approach has a drawback because unmodelled aspects of the system are often left out, causing the model to diverge from the real system. In contrast, a black box is built using data collected from experiments with the real system. One disadvantage of this approach is that it is hard to obtain a generalized model that works in different scenarios since it does not follow any physics representation \cite{schoener2019global}. Recent efforts utilizing deep learning in system identification show promising results \cite{forgione2021continuous,xu2022physics,chee2022knode}. 

In this work, we provide a neural network architecture to accurately model the dynamics of an ASV, we use the learned model to implement MPC to solve the station keeping problem, and we discuss its applicability to underwater docking. We use a neural network model based on \cite{forgione2021continuous} and trained on simulation data, as well as field data collected during deployments at Fairfield Lakes and Lake Harner in Tippecanoe County, IN. This method takes advantage of some prior knowledge of the real system and considers actual data which allows a better match between the model and the real system. Additionally, we compare the performance of the Neural Network Simulation Error Minimization Model Predictive Controller (NNSEM-MPC) against other nonlinear and data-driven controllers in simulation.
Furthermore, we showcase the viability of the NNSEM-MPC controller through successive stages: initially, a proof of concept demonstration on the portable and cost-effective custom ASV named BREAM (Boat for Robotic Engineering and Applied Machine Learning), followed by a refined deployment on the 16' Wave Adaptive Modular Vehicle (WAM-V 16). While this latter offers an improved sensor package, its transportation and deployment entail a more complicated process. Both of the experimental validation stages are assessed for linear and rotational speed stability which provides the best window of opportunity for an AUV to perform terminal homing under wind disturbances and successfully dock into the docking station installed on the ASV.  



The rest of the paper is organized as follows: In Section~\ref{ssec:sk_math_model} the station keeping problem statement is defined along with a 3-DOF reference model of an ASV. In section~\ref{ssec:sk_nn_model} the proposed NNSEM-MPC architecture is presented, followed by a description of the baseline controllers that are compared to the proposed controller in section ~\ref{ssec:ref_controllers}. In Section~\ref{sec:results} we provide the details of the simulation validation and describe the proof of concept implementation process on custom-made BREAM-ASV, and 
 validation on the commercial off-the-shelf WAM-V 16 ASV. Finally, conclusions and future work are discussed in Section~\ref{sec:conc}. 

\section{Methodology}\label{sec:methodology}
\subsection{Station Keeping Model} \label{ssec:sk_math_model}
\begin{figure}[t]
\vspace{0.3cm}
\centering
\includegraphics[width=0.5\columnwidth]{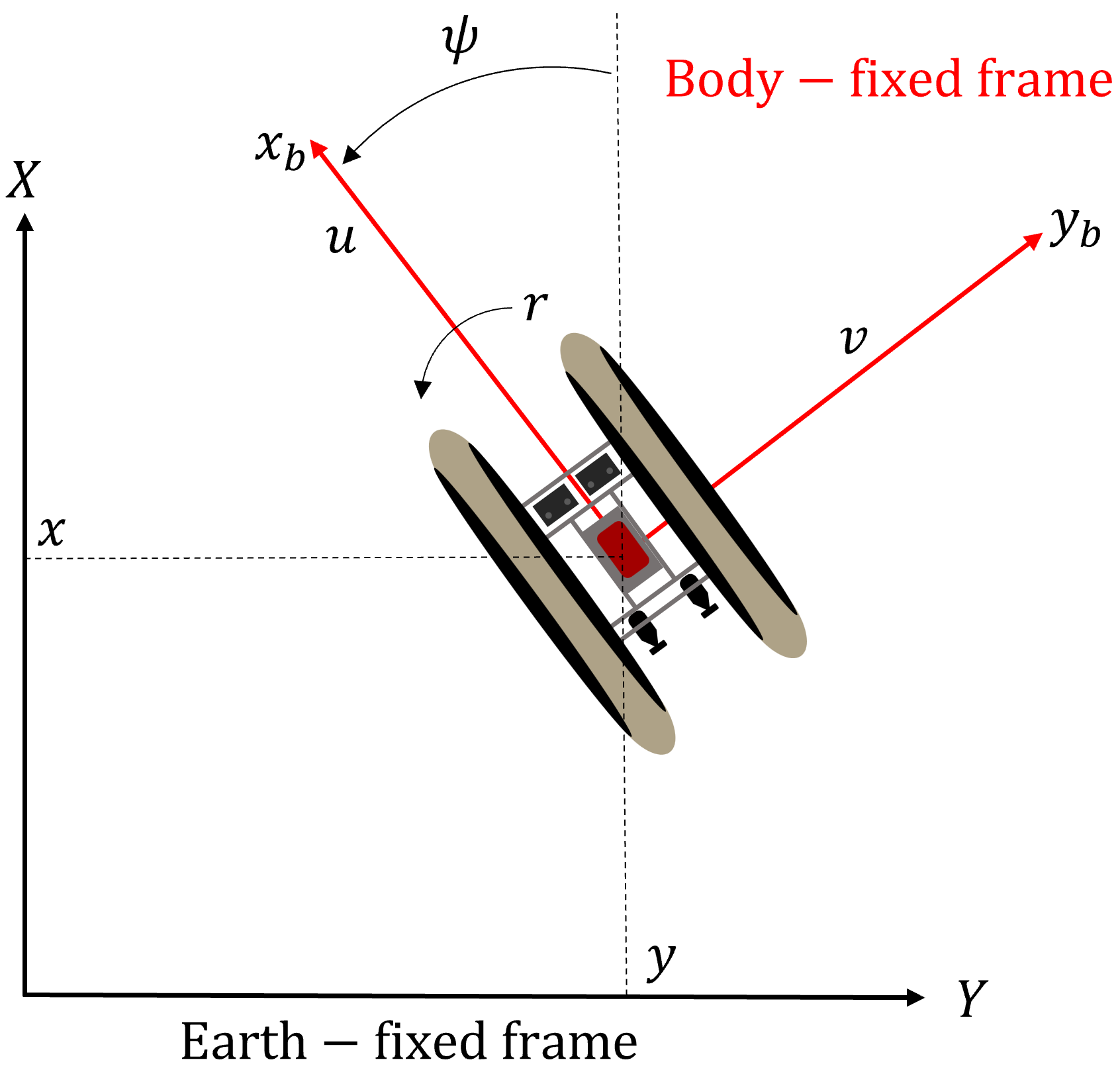}
\caption{Earth fixed frame and body fixed frame definition for the 3DOF dynamical model of the BREAM-ASV.}
\label{fig:eff_bff_iver}
\end{figure}

The ASVs used for this work are differential vehicles propelled via two motors. The motors are fixed to the vehicle, and therefore the turning motion is achieved by driving each of the motors at a different speed. To model the dynamics of these vehicles we consider that gravity and buoyancy generate restoring forces that cancel out the pitch, roll, and heave motions, reducing the final dynamics of the vehicle to a 3-DOF model considering only the surge, sway, and yaw motions as shown in Fig.~\ref{fig:eff_bff_iver}. As described in \cite{fossen2011handbook,qu2022nonlinear,sarda2016station} the 3-DOF equations of motions describing the simplified dynamics (SD) of an ASV can be expressed as follows:


\begin{align}
    \pmb{\dot{\eta}} = \pmb{R}(\psi)\pmb{v}\\
    \pmb{M}\pmb{\dot{v}} + \pmb{C}(\pmb{v})\pmb{v} + \pmb{D}(\pmb{v})\pmb{v} = \pmb{\tau} + \pmb{\tau}_w \label{eq:asv_dyn_model}
\end{align}

where $\pmb{\eta} = [x,y,\psi]^{T}$ consists of the $(x,y)$ position and the orientation $\psi$ expressed in NED coordinates. The speed vector $\pmb{v}=[u,v,r]^{T}$ consists of the linear velocities $(u,v)$ in the surge and sway directions respectively, and the rotation velocity $r$ in the body fixed frame as shown in Fig.~\ref{fig:eff_bff_iver}. The thrust vector $\pmb{\tau}=[K(u^p + u^s),0,\frac{B}{2}(u^p - u^s)]^T$ where $K$ is the thrust coefficient, and $B$ is the beam of the vehicle contains the force produced by the thrust of the port and starboard trolling motors $\pmb{u}^{thrust}=[u^p,u^s]^T$. The wind disturbance $\pmb{\tau}_w=[\tau^u_w,\tau_w^v,\tau_w^r]^T$ produced by the wind speed $\pmb{w}=[w^u,w^v]$ is included to consider the impact in the equations of motions to the surge and sway directions of the ASV. The matrix $\pmb{R}(\psi)$ is the rotation matrix used to transform from the body fixed frame to the earth fixed frame. 
The mass matrix $\pmb{M}$ includes the rigid body as well as the added mass components and can be expressed as:

\begin{align}
 \pmb{M}
 =
  \begin{bmatrix}
    m -X_{\dot{u}} &  0    &  0     \\
    0  &  m-Y_{\dot{v}}    &  mx_F-Y_{\dot{r}}  \\
    0  &  mx_G-N_{\dot{v}} &  Iz-N_{\dot{r}}    \\
  \end{bmatrix}
\end{align}

The Coriolis matrix $\pmb{C}(\pmb{v})=\pmb{C}_{RB}(\pmb{v}) + \pmb{C}_{A}(\pmb{v})$ is composed of the rigid body and hydrodynamic components.

\begin{align}
\setstacktabbedgap{0pt}
 \pmb{C}_{RB}(\pmb{v})
 =
  \begin{bmatrix}
    0            &  0          &  -m(x_{G}r+v)  \\
    0            &  0          &  -m(y_{G}r-u)  \\
    m(x_{G}r+v)  & m(y_{G}r-u) &  0             \\
  \end{bmatrix},\\
  \pmb{C}_{A}(\pmb{v})
 =
  \begin{bmatrix}
    0            &  0           &  Y_{\dot{v}}v+(\frac{Y_{\dot{r}}+N_{\dot{v}}}{2})r  \\
    0            &  0           &  -X_{\dot{u}}u  \\
    -Y_{\dot{v}}v-(\frac{Y_{\dot{r}}+N_{\dot{v}}}{2})r  & X_{\dot{u}}u &  0             \\
  \end{bmatrix}
\end{align}

Finally, the damping matrix $\pmb{D}(\pmb{v})=\pmb{D}_{NL}(\pmb{v})+\pmb{D}_{L}$ where $\pmb{D}_{L}$ is the linear drag term $\pmb{D}_{NL}(\pmb{v})$ is the nonlinear drag term. The damping matrix is mainly dominated by the linear components, which is a common assumption for maneuvering like station keeping.

\begin{align}
\setstacktabbedgap{0pt}
 \pmb{D}_L
 =
  \begin{bmatrix}
    X_{u}    &  0       &  0      \\
    0        &  Y_{v}   &  0      \\
    0        &  0       &  N_{r}  \\
  \end{bmatrix},\\
 \pmb{D}_{NL}(\pmb{v})=\hspace*{6cm}\\\nonumber
  \begin{bmatrix}
    X_{u|u|}|u|    &  0       &  0      \\
    0        &  Y_{v|v|}|v|+Y_{v|r|}|r|   &  Y_{r|v|}|v|+Y_{r|r|}|r|      \\
    0        &  N_{v|v|}|v|+N_{v|r|}|r|       &  N_{r|v|}|v|+N_{r|r|}|r|  \\
  \end{bmatrix}  
\end{align}

\subsection{Neural Network Model Predictive Controller} \label{ssec:sk_nn_model}
 This work proposes an NNSEM-MPC for ASV station keeping. This approach adopted the truncated simulation error minimization approach described in \cite{forgione2021continuous} to train a neural network to predict the ASV states based on control inputs, previous states, and disturbances. Then this neural network is used by the MPC to explore future states and generate optimal control commands.
 
 The truncated simulation error minimization method processes batches of the collected dataset containing $q\in\mathbb{N}$ subsequences of length $m\in\mathbb{N}$. Considering that all the states can be measured, including the disturbances produced by the wind, the neural network to model the ASV can be written as follows:

\begin{align}
\hat{\dot{\pmb{v}}} = \mathcal{N}_f(\hat{\pmb{v}},\pmb{u}^{thrust},\pmb{w};\theta)\\\nonumber
\hat{\pmb{v}}(0) = \pmb{v}_0
\end{align}

where $\mathcal{N}_f$ represents a 2-layer fully connected, feedforward neural network with parameters $\theta$ and one hidden layer of size $n_h$. Such a network estimates the derivative of system state $\hat{\dot{\pmb{v}}}$, given the known initial condition $\pmb{v}_0$, the estimated state $\hat{\pmb{v}}$, the measured wind speed $\pmb{w}$, and the control input $\pmb{u}^{thrust}$. 

The goal of the network is to learn a model able to predict a sequence of length $m\geq T$, where $T$ is the MPC time horizon. For that, the neural network output is transformed into the earth fixed frame using the rotation matrix $\pmb{R}(\psi)$, and then integrated using a forward Euler scheme with constant step size $\Delta t$. Then for a time $t$ computing a one-step ahead prediction can be written as follows:

\begin{align}
\hat{\pmb{v}}_{t+1} = \hat{\pmb{v}}_{t} + \Delta t \cdot\mathcal{N}_f(\hat{\pmb{v}}_t,\pmb{u}^{thrust}_t,\pmb{w}_t;\theta); \
\hat{\pmb{v}}(0) = \pmb{v}_0\\\nonumber
\hat{\pmb{\eta}}_{t+1} = \hat{\pmb{\eta}}_{t} + \Delta t \cdot\hat{\pmb{v}}_{t}; \hat{\pmb{\eta}}(0) = \pmb{\eta}_0
\end{align}

\begin{figure}[t]
\vspace{0.3cm}
\centering
\includegraphics[width=0.6\columnwidth]{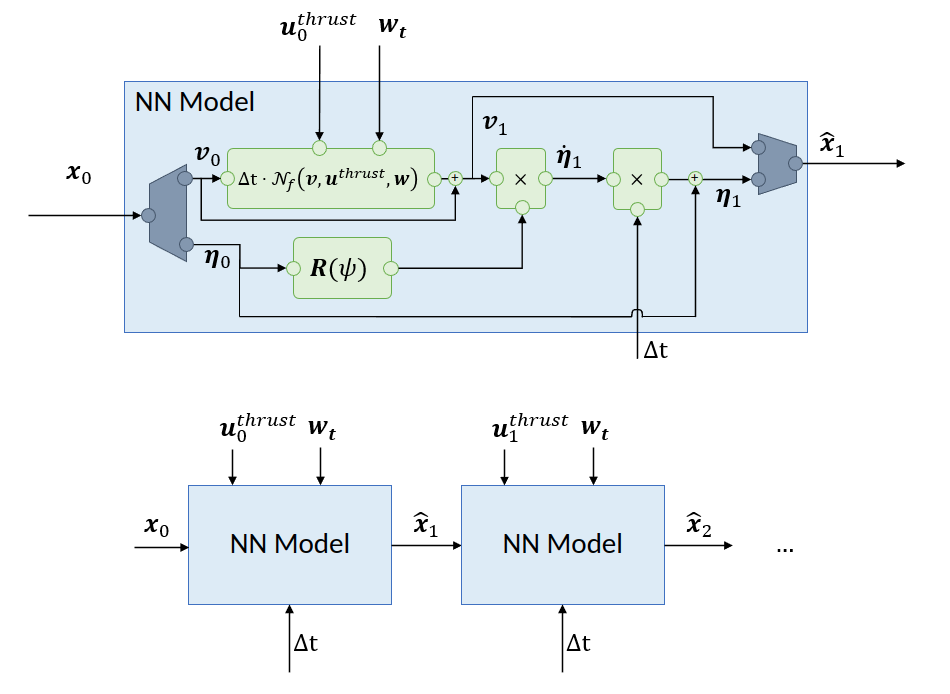}
\caption{Neural network structure using forward Euler integration with step size $\Delta t$ to estimate a sequence of $m$ data points. }
\label{fig:sk_nn_arch}
\end{figure}

At the end of the neural network model, the position and the output of the neural network are concatenated to obtain the final estimated state $\hat{\pmb{x}}_t=[\hat{\pmb{v}}^T_t,\hat{\pmb{\eta}}^T_t]^T$ as shown in Fig.~\ref{fig:sk_nn_arch}.

To train the neural network we used a mean squared error approach over the whole sequence $m$ for each sample in the batch of length $q$. The aim is to penalize the distance between the predicted sequence $\hat{\pmb{v}}_{0:m-1}$ and the measured sequence $ \pmb{v}_{0:m-1}$ defined as:

\begin{gather}
J(\theta,\pmb{v},\pmb{u}^{thrust},\pmb{w}) = 
\frac{1}{qm}\sum_{j=0}^{q-1}\sum_{t=0}^{m-1}\| \pmb{v}_{j,t} - \hat{\pmb{v}}_{j,t} \|^2
\end{gather}

\begin{figure}[t]
\vspace{0.3cm}
\centering
\includegraphics[width=0.5\columnwidth]{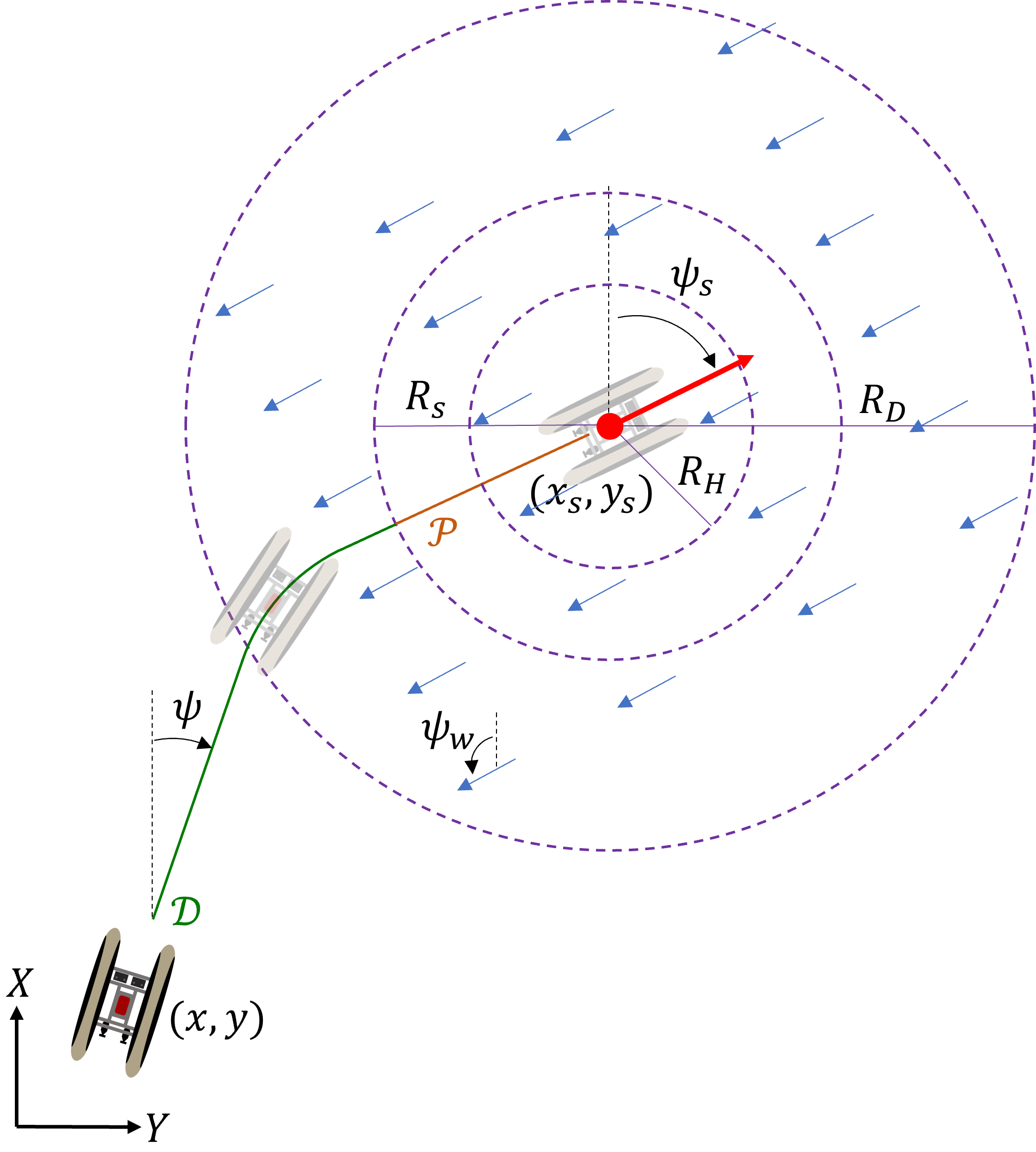}
\caption{Station keeping procedure of the ASV to achieve the final configuration $\pmb{\eta}_s = [x_s,y_s,\psi_s]^T$ given a wind disturbance with direction $\psi_w$. The ASV will start following a Dubins path $\mathcal{D}$, and then it will be driven by the MPC controller following a trajectory generated by the MPC controller $\mathcal{P}$. The parameters $R_S$ and $R_D$ can be adjusted to improve the station keeping procedure.}
\label{fig:asv_sk_setup}
\end{figure}

MPC is a type of process control that involves finding a sequence of actions given a set of optimization constraints, to achieve the desired goal. This is of particular use when the mathematical model of the system is available, and a cost function can be defined to optimize the system behavior \cite{amos2018differentiable}. This work utilizes MPC as a way to optimally drive an underactuated ASV to a target configuration $\pmb{\eta}_s = [x_s,y_s,\psi_s]^T$ while rejecting environmental disturbances. Throughout this process, the ASV approaches $\pmb{\eta}_s$ using a Dubins path planner as the one used in \cite{page2022path} and \cite{page2021underwater} to create a prescribed path $\mathcal{D}$, such a path is followed using an integral line of sight (ILOS) reference tracker. Once the ASV is close to the target location $\sqrt{(x-x_s)^2+(y-y_s)^2}<R_S$ the MPC controller takes over the control of the vehicle to find a sequence of control outputs $\pmb{u}^{thrust}$, and smoothly drive the vehicle through an MPC generated path $\mathcal{P}$ towards the position $(x_s,y_s)$. As the ASVs used in this work are underactuated, keeping position and orientation simultaneously is a complicated task; therefore, an additional circular region of radius $R_H$ is defined. Within this region, the focus of the MPC controller is to keep heading and the forward distance to the goal. If at some point the vehicle exits the region bounded by $R_H$, the focus of MPC is dynamically adjusted to put more priority on keeping position without considering the heading. The target heading $\psi_s$ is manually selected to be along, against, or perpendicular to the wind direction. It allows us to better study the effect of the disturbance and compare the results. If at some point while keeping station with the MPC controller, the vehicle is far enough from the target location $\sqrt{(x-x_s)^2+(y-y_s)^2}>R_D$ the Dubins planner and the ILOS controller take control of the vehicle and start the process over again. In the experimental valications, $R_S$ is $15 m$ and $R_S$ is $20 m$. In this way, Dubin's planner keeps the ASV in a range where the learned model works better, resulting in a better performance of the MPC controller. Fig.~\ref{fig:asv_sk_setup} illustrates the aforementioned station keeping procedure.

MPC as a model-based technique requires an accurate way to determine the system's state based on calculated outputs within a time horizon window $T$. To have better control over the system states, the state space of the ASV is defined to include velocity, and position $\pmb{x}_t = [\pmb{v}^T_t,\pmb{\eta}^T_t]^T\in \mathcal{X}$. The future relative wind velocity cannot be measured, thus we assume the wind speed is constant in the global frame during prediction. We can obtain the relative wind velocity by multiplying the initial wind speed by a transformation matrix using the predicted heading. The control output $\pmb{u}^{thrust}\in\mathcal{U}$ corresponds to the thrust generated by the two motors. In this way, MPC is configured to solve the following optimization problem:

\begin{gather}
 \pmb{x}_{0:T-1},\pmb{u}^{thrust}_{0:T-1} = \argmin_{\pmb{x}_{0:T-1}\in\mathcal{X},\pmb{u}^{thrust}_{0:T-1}\in\mathcal{U}} \sum^{T-1}_{t=0} C_t(\pmb{x}_t,\pmb{u}^{thrust}_t,\pmb{w}_t)\\\nonumber
 \subto \pmb{x}_{t+1} = f(\pmb{x}_t,\pmb{u}^{thrust}_t,\pmb{w}_t), \pmb{x}_0 = \pmb{x}_{init}
\end{gather}

where the cost function is defined in terms of the augmented state vector $\pmb{\alpha}_t = [\pmb{x}_t^T,{\pmb{u}^{thrust}_t}^T]^T$, the augmented goal state $\pmb{\alpha}^{*} = [{\pmb{x}^*}^T,\pmb{0}]^T$, and the goal weight vector $\pmb{g}_w=[k_u,k_v,k_r,k_x,k_y,k_\psi]^T$ as follows:

\begin{gather}
C_t(\pmb{x}_t,\pmb{u}^{thrust}_t) = \frac{1}{2}\pmb{\alpha}_{t}^{T}\pmb{D}(\pmb{g}_w)\pmb{\alpha}_t-(\sqrt{\pmb{g}_w}\circ\pmb{\alpha}^{*})^{T}\pmb{\alpha}_t
\end{gather}

where $\pmb{D}(\pmb{g}_w)$ is a diagonal matrix including the goal weights and can be adjusted to achieve the desired behavior of the system, and also to penalize large control actions. $\circ$ is the pointwise product operator, and $\sqrt{\pmb{g}_w}$ computes the square root of the elements in $\pmb{g}_w$. With the optimization problem defined, the procedure can be repeated to obtain the optimal control output over the time horizon. 
The optimizer utilizes a model to encounter an optimal control output $\pmb{\tau}_t$ given the available model $\mathcal{N}_f$ and the predefined set of weights $\pmb{g}_w$.


The optimization problem was first solved by \verb|mpc.torch| \cite{amos2018differentiable}, a differentiable MPC solver for PyTorch. This library requires a linearized dynamics function at each time step. Then, the library solves the MPC problem formulated as a linear quadratic regulator (LQR) problem in a forward pass. This library was initially selected due to the ease of integration with the neural network model; however, this library was replaced in the refined implementation with a faster module that allows the usage of trained torch models with CasADi.


\subsection{Baseline Controllers} \label{ssec:ref_controllers}
 
  This section describes the different controllers used as a baseline to compare the performance of the proposed NNSEM-MPC approach.


\subsubsection{Backstepping Controller}

Backstepping control combines the choice of a Lyapunov function with a feedback controller \cite{VAIDYANATHAN20211}. This results in a stable and robust system; however, according to the simulation and experimental results reported in \cite{sarda2016station}, a regular PD controller could not perform due to unmodelled dynamics and external disturbances. The nonlinear P-D backstepping controller implementation of this controller uses equations of motions defined in Eq. \ref{eq:asv_dyn_model} with the simplifications described in \cite{sarda2016station}. The Lyapunov exponent gain matrix, $\pmb{\Lambda} = \diag\{\Lambda_1,\Lambda_2,\Lambda_3\}$, was defined according to \cite{sarda2016station}, which took into consideration different physical and environmental factors. The same matrix is used in the implementation of the controller, for which a reference trajectory is defined as $\dot{\pmb{\eta}}_r=\dot{\pmb{\eta}}_s-\pmb{\Lambda} \pmb{\eta}_e$, where $\pmb{\eta}_e$ is the earth-fixed tracking error, $\dot{\pmb{\eta}}_s$ is the time derivative of the desired configuration. To have a measure of tracking another a tracking surface is defined in terms of the Lyapunov exponents using $
\pmb{s}=\dot{\pmb{\eta}}_e+\pmb{\Lambda} \pmb{\eta}_e$.
Finally, a feedback control law $\pmb{\tau}$ to find the corrective torques and forces needed in each axis is defined according to the equations below.

\begin{equation}
    \begin{aligned}
        \pmb{\tau} & =\pmb{M}_1\left[\pmb{R}(\psi)^T \ddot{\pmb{\eta}}_r+\pmb{R}(\dot{\pmb{\psi}})^T
        \dot{\pmb{\eta}}_r\right]+\pmb{C}_{1}(\pmb{v}) \pmb{R}(\psi)^T \dot{\pmb{\eta}}_r \\ 
        &  + \pmb{D}_{1} \pmb{R}(\psi)^T \dot{\pmb{\eta}}_r-\pmb{R}(\psi)^T \pmb{K}_d \pmb{s}-\pmb{R}(\psi)^T \pmb{K}_p \pmb{\eta}_e
    \end{aligned}
\end{equation}

Where, $\pmb{M}_{1}, \pmb{C}_1(\pmb{v})$ and $\pmb{D}_1 $
are the simplified versions of the mass, Coriolis, and damping matrices according to the assumptions mentioned in \cite{sarda2016station}.


\subsubsection{Sliding Mode Controller}
Sliding mode controller has proven effective to improve the robustness to slowly varying environmental factors like the tidal currents \cite{sarda2016station}. It does this by breaking the nonlinear behavior of the system, and applying chattered signals which are not continuous \cite{761053}. The system is driven towards the sliding surface and once it intercepts it, a very high gain is applied which tries to force the system to follow the surface. Once the sliding surface is reached, the system becomes robust to some of the disturbances \cite{VAIDYANATHAN20211}.
One of the flaws in this design method is the possibility of damage to actuators due to chattering \cite{MUNOZ201761}.
A sliding surface is defined as $
\pmb{s}=\dot{\pmb{\eta}}_e+2 \pmb{\Lambda} \pmb{\eta}_e+\pmb{\Lambda}^2 \int_0^t \pmb{\eta}_e d t
$. Further, the reference trajectory is redefined for this controller as $
\dot{\pmb{\eta}}_r=\dot{\pmb{\eta}}_s-2 \pmb{\Lambda} \pmb{\eta}_e-\pmb{\Lambda}^2 \int_0^t \pmb{\eta}_e d t
$. Finally, the control law for finding corrective thrust action $\pmb{\tau}$ is defined as follows.

\begin{equation}
    \begin{aligned}
        \pmb{\tau}= &\pmb{M}_{1}\left[\pmb{R}(\psi)^T \ddot{\pmb{\eta}}_s+ \pmb{R}(\dot{\psi})^T \dot{\pmb{\eta}}_s\right]+\pmb{C}_{\mathbf{1}}(\pmb{v}) \pmb{R}(\psi)^T \dot{\pmb{\eta}}_s+ \\ &
        \pmb{D}_{\mathbf{1}}(\pmb{v}) \pmb{R}(\psi)^T \dot{\pmb{\eta}}_s-\pmb{R}(\psi)^T \pmb{U} \cdot \operatorname{sat}\left(\pmb{E}^{-1} \cdot \pmb{s}\right)
    \end{aligned}
\end{equation}

In this control law, $\pmb{U}$ is defined as a diagonal matrix with all positive elements which defines the bound on uncertainties. The thickness of the boundary layer within which the system will ``slide" is defined by the vector $\pmb{E}$.

\subsubsection{KNODE-MPC}
The knowledge-based neural ordinary differential equations model predictive controller (KNODE-MPC) is based on the NODE approach, which has been widely used to approximate different dynamical systems as discussed in \cite{forgione2021continuous}. NODE uses the backward time integration of the adjoint sensitivities to differentiate through the ordinary differential equation solution \cite{NEURIPS2018_69386f6b}. KNODE is an extension of NODE that couples the physics knowledge of the system (if available) with neural networks and can be written as follows:

\begin{align}
\dot{\pmb{x}} = g_{\lambda}(\tilde{f}(\pmb{x},\pmb{u},\pmb{w}),f_{\theta}((\pmb{x},\pmb{u},\pmb{w};\theta);\lambda)
\end{align}

Where $\tilde{f}$ represents the prior knowledge of the system, which can be imperfect. $f_{\theta}$ is a neural network parameterized with $\theta$. Finally, $\tilde{f}$ and $f_{\theta}$ are coupled using a neural network $g_{\lambda}$ that is co-trained with $f_{\theta}$, and its objective is to learn the parameters $\lambda$ that best reduce the approximation error. In combination with a model predictive framework, the controller KNODE-MPC can be used to generate a sequence of optimal control inputs in the time horizon $T$ to achieve the station keeping task.
\begin{figure}[t]
\vspace{0.3cm}
\centering
\includegraphics[width=0.5\columnwidth]{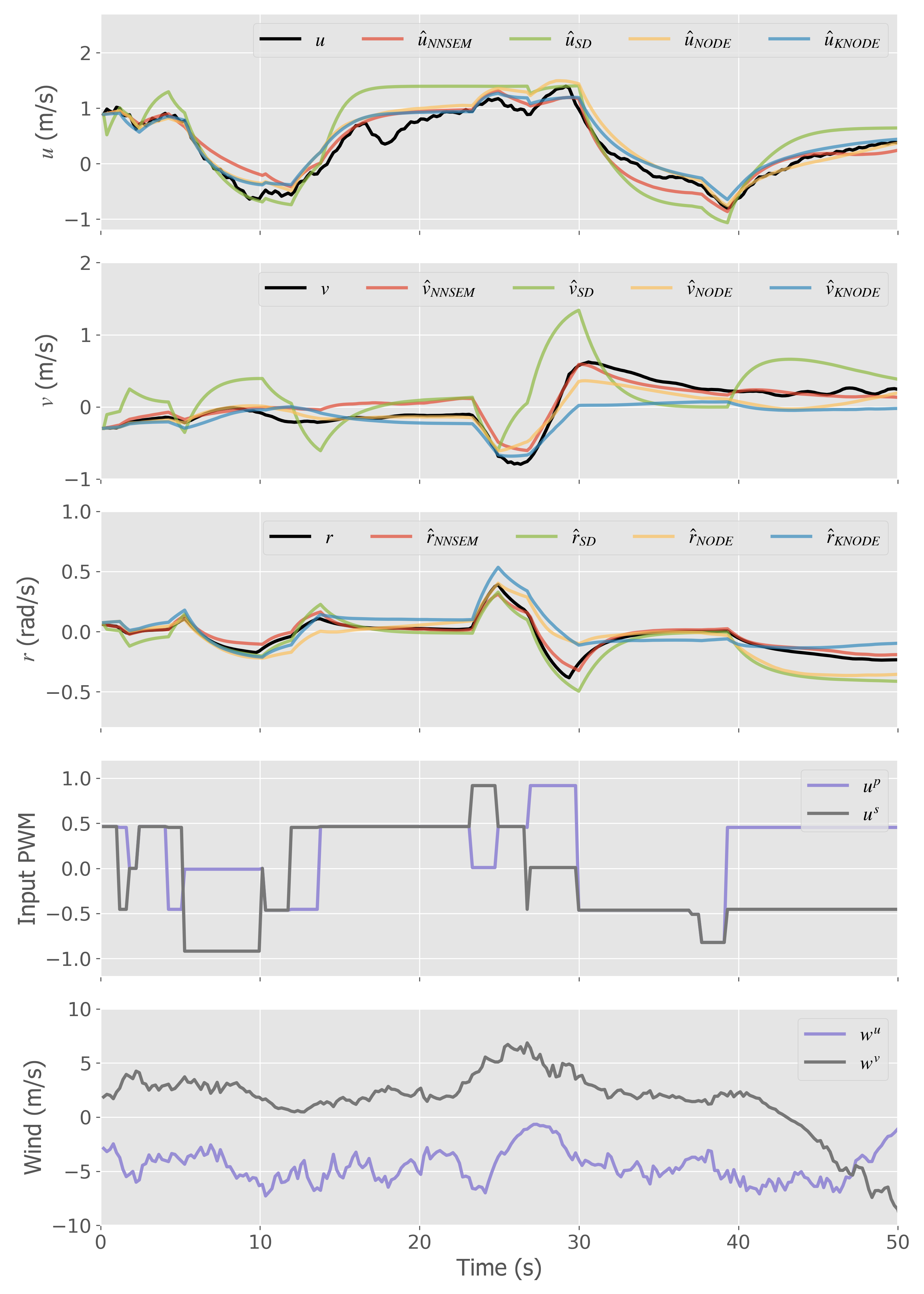}
\caption{Measured output $u$,$v$,$r$ (black) and model simulation $\hat{u}$,$\hat{v}$,$\hat{r}$ obtained by NNSEM model (red), SD model (green), NODE model (yellow), and KNODE model (blue). Input PWM curves are the control signals. Wind curves are the wind speed measured from the robot frame.}
\label{fig:system_id}
\vspace{-0.3cm}
\end{figure}
\section{Results} \label{sec:results}
\subsection{Data Collection}
To train the neural network model, a series of maneuvers are defined to cover the most values for the states of the system. For this, ASV is programmed to perform a predefined set of actions in addition to manual driving to allow for a repeatable data collection process. The predefined maneuvers include linear motions in forward and backward directions, circular maneuvers, and zigzagging at different velocities. The manual driving maneuvers are station keeping control attempts using joysticks by an operator. The ASV is equipped with a left and a right trolling motors, which accept a Pulse Width Modulated (PWM) signal as the input ($1$ being maximum forward thrust and $-1$ being maximum reverse thrust). Data is collected at $5$ Hz for both simulation and real-world tests. A total length of $600$ seconds of data is collected for simulation and a total length of $500$ seconds of data is collected for real-world tests. Different datasets are collected for various wind conditions to allow better generalization of the neural network model. The collected data is divided into training and test datasets to train and evaluate the performance of the learned model.

To estimate the speed, an accelerometer is used; this approach tends to drift over time, which is solved using a finite impulse response filter for the acceleration and fusing the speed calculated from the integration of the acceleration with GPS measurements. This process allows stable and accurate speed measurements, resulting in better state estimation, and thus, a better NNSEM-MPC performance. Additionally, an anemometer is used to measure the disturbances. The anemometer provides 16 different analog outputs for the wind direction and a square wave output for the wind speed. These signals are processed using an onboard Arduino board. The main processor receives the resulting wind speed and direction through serial communication.

\renewcommand{\tabcolsep}{3.0pt}
\begin{table}[t]
\centering
\vspace{0.3cm}
\begin{tabular}{lllll}
\hline
\multirow{2}{*}{\textbf{Model} } & \multicolumn{3}{l}{\textbf{velocity MAE}} & \multirow{2}{*}{\textbf{\begin{tabular}[c]{@{}l@{}}Iteration\\Time (s)\end{tabular}}} \ \\ \cline{2-4}
                   &$\boldsymbol{u}$\textbf{(m/s)}   &$\boldsymbol{v}\textbf{(m/s)}$      &$\boldsymbol{r}\textbf{(rad/s)}$      &                    \\ \hline
\textbf{NNSEM} & 0.14                                           & 0.09                                         & 0.02                                          & 0.68                                         \\
\textbf{SD} & 0.32                                           & 0.31                                         & 0.07                                          & 2.38                                         \\
\textbf{NODE} & 0.15                                           & 0.10                                         & 0.07                                            & 1.06                                         \\
\textbf{KNODE} & 0.13                                           & 0.17                                         & 0.07                                           & 2.50                                         \\\hline
\end{tabular}
 
\caption{State estimation results and iteration time comparison between NNSEM model, SD model, NODE model, and KNODE model.
\label{tbl:sim_result}}
\end{table}

\subsection{Neural Network Model Training and Evaluation}



To train a neural network to estimate the system dynamics, the step size $\Delta t$ is set to be $0.2$ s for the simulation. The model is created with a single hidden layer $n_h=12$, and trained for 20000 epochs with a batch size of $q=64$, a prediction window $m=20$, and a learning rate of $0.001$. To prevent any unbalance in the prediction of the states, all of the input, disturbance, and state are normalized on $[-1,1]$ before being fed into the neural network. This normalization is tightly coupled to the expected operating range of the station keeping controller. Training is done on a desktop computer with a GeForce 2080 GPU and an Intel Core i7-8700 CPU. 

To test the performance of the trained models, we use the model to generate velocity predictions on the test set. The mean wind speed is 6 m/s. As shown in Fig.~\ref{fig:system_id}, only giving initial state, control inputs, and wind disturbances, the NNSEM model is able to predict future states for $50$ s ($250$ steps) with the mean absolute error (MAE) of $0.14$m/s, $0.09$m/s, and $0.02$ rad/s for $u$, $v$, and $r$. The error for $u$ is only $0.1$m/s larger than KNODE, while the error for $v$ and $r$ are the smallest among all the controllers. We hypothesize that the difference in the results is primarily due to the weight update optimization method, which in the NODE-based methods is done through adjoint sensitivities rather than back-propagation. Finally, KNODE approach is sensitive to the prior knowledge of the system, which can be observed in the MAE for $v$ and $r$ where the SD exhibits the largest error, making it hard for the adaptive part of KNODE to estimate these states. We apply the NNSEM approach to the BREAM-ASV and WAM-V 16 with real-world data. After the training, the mean absolute error for $u$, $v$, and $r$ is $0.21$m/s, $0.18$m/s, and $0.12$ rad/s on a test set that lasts for $50$ s for BREAM-ASV and $0.22$m/s, $0.10$m/s, and $0.13$ rad/s on a test set that lasts for $20$ s for WAM-V 16. Due to the sensor noise of the wave disturbance, the NNSEM model's prediction error in the real-world dataset is larger than the error in the simulation dataset but still acceptable.


\subsection{Validation in Simulation}
\label{ssec:nn_eval}

\begin{figure}[t]
\centering
\vspace{0.2cm}
\includegraphics[width=0.7\columnwidth]{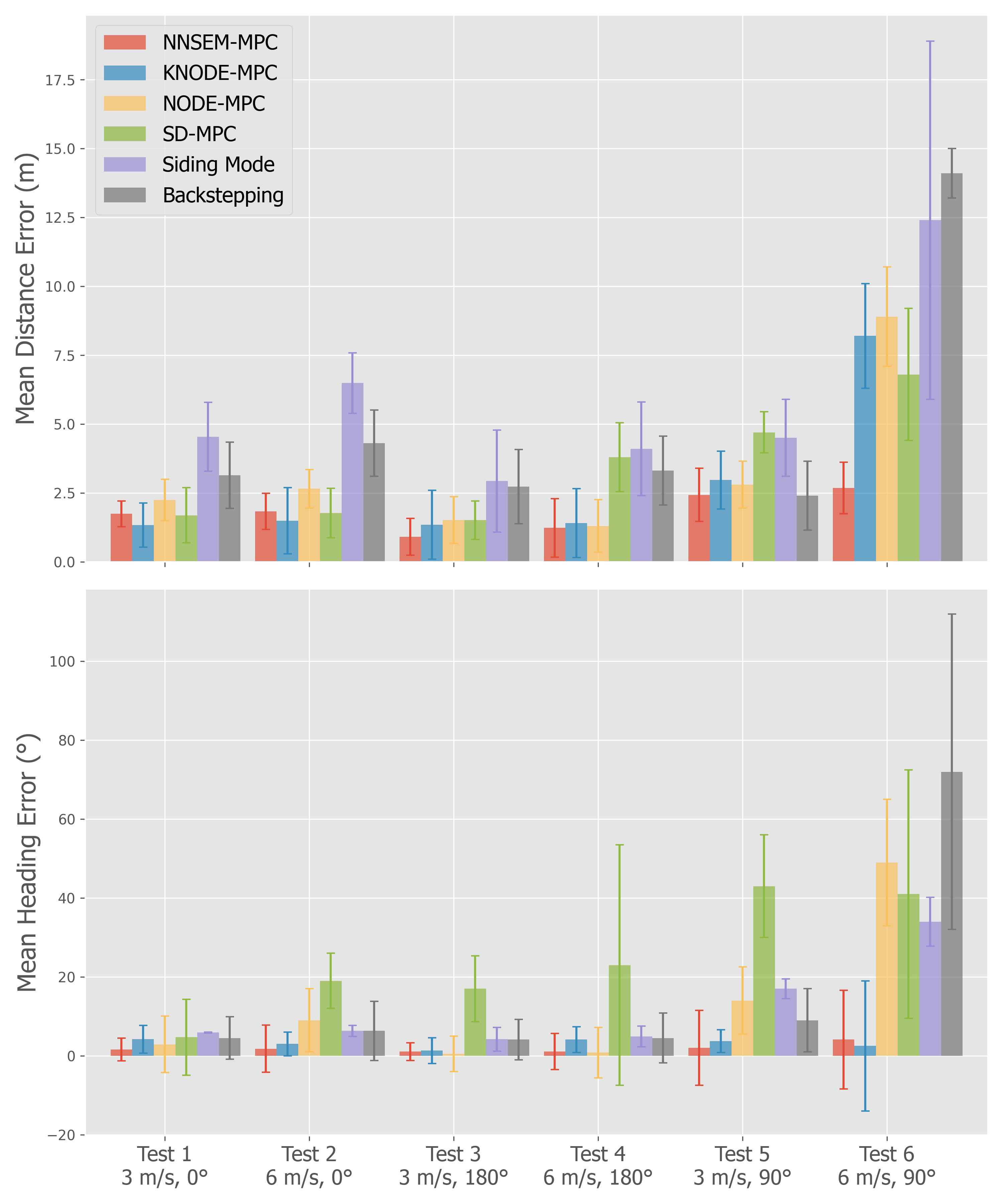}
\caption{Comparison of the controllers for 6 tests in simulation. The target heading is $0$° for all tests. The peaks of the bars represent the mean and the ends of the error bars depict $\pm 0.5$ standard error. The wind condition in the global frame is indicated under the test number.}
\label{fig:sim_compare}
\vspace{-0.3cm}
\end{figure}
To compare the performance of the NNSEM-MPC with backstepping, sliding mode controllers, NODE-MPC, and KNODE-MPC, six tests were carried out in the Gazebo simulator using different wind speeds and wind directions. The simulation validation is also done on the desktop computer with a GeForce 2080 GPU and an Intel Core i7-8700 CPU. As shown in Fig. \ref{fig:sim_compare}. Tests $1$, $3$, and $5$ use a mild wind disturbance while tests $2$, $4$, and $6$ use a stronger disturbance. For all tests, the starting position and target pose are the same. The boat starts from $(-14,-3)$ m in the global frame. The goal position is $(0,0)$ m, and the goal heading is $0$ degree. Therefore, for tests $1$, $2$, the goal heading is along the wind, test $3$, $4$ is against the wind, and test $5$, $6$ perpendicular to the wind. To ensure a fair comparison, all MPC related controllers use the \verb|mpc.torch| solver with the same optimization parameters. The number of LQR iterations to perform is set to $2$. All model linearization is done by PyTorch’s automatic differentiation. The penalty term applied to control change is set to $0.001$. The time horizon is set to $T=4$ s. As shown in Table \ref{tbl:sim_result}, the mean iteration times of the first $10$ steps for NNSEM-MPC, SD-MPC, NODE-MPC, and KNODE-MPC are $0.68$ s, $2.38$ s, $1.06$ s, and $2.50$ s respectively. NNSEM-MPC has the shortest iteration time due to the simplicity of the model, and KNODE-MPC has the largest iteration time because it uses both an SD model and a NODE model. The simulation time is adjusted based on the iteration time so that the step time is $0.2$ s.

The results of the tests are summarised in Fig. \ref{fig:sim_compare}. The performance of the station keeping strategy is evaluated not only using the magnitude of the distance and the heading errors ($e_d,e_h$) but also considering a steady behavior. For that reason, the standard deviation of the distance $s_d$ and heading $s_h$ are also considered in the overall comparison. In this way, we select the best controller for each test case by getting the lowest penalty score $ps = e_d + s_d + |e_h| + s_h$. Following this approach, NNSEM-MPC achieves better performance in test cases 1,3,4, and 6, while KNODE-MPC reaches the best performance in test cases 2 and 5. Overall, NNSEM-MPC achieves a better performance than the rest of the controllers when controlling an underactuated ASV to reach a goal and keep station. When the ASV's target heading is against the wind, it obtains the lowest error and penalty score in both mild and strong wind scenarios. In scenarios where the target heading is perpendicular to the wind, the distance error and the heading error are the largest. This is mainly because, in these cases, the lateral error keeps increasing, and the underactuated ASV is not able to counteract this without having to replan a path to the target, while in other cases, the wind will not cause the lateral error to increase, and the controller only needs to eliminate the heading error and longitudinal error.

\subsection{Experimental Test Setup}
\begin{figure}[t]
\vspace{0.3cm}
\centering
\includegraphics[width=0.9\columnwidth]{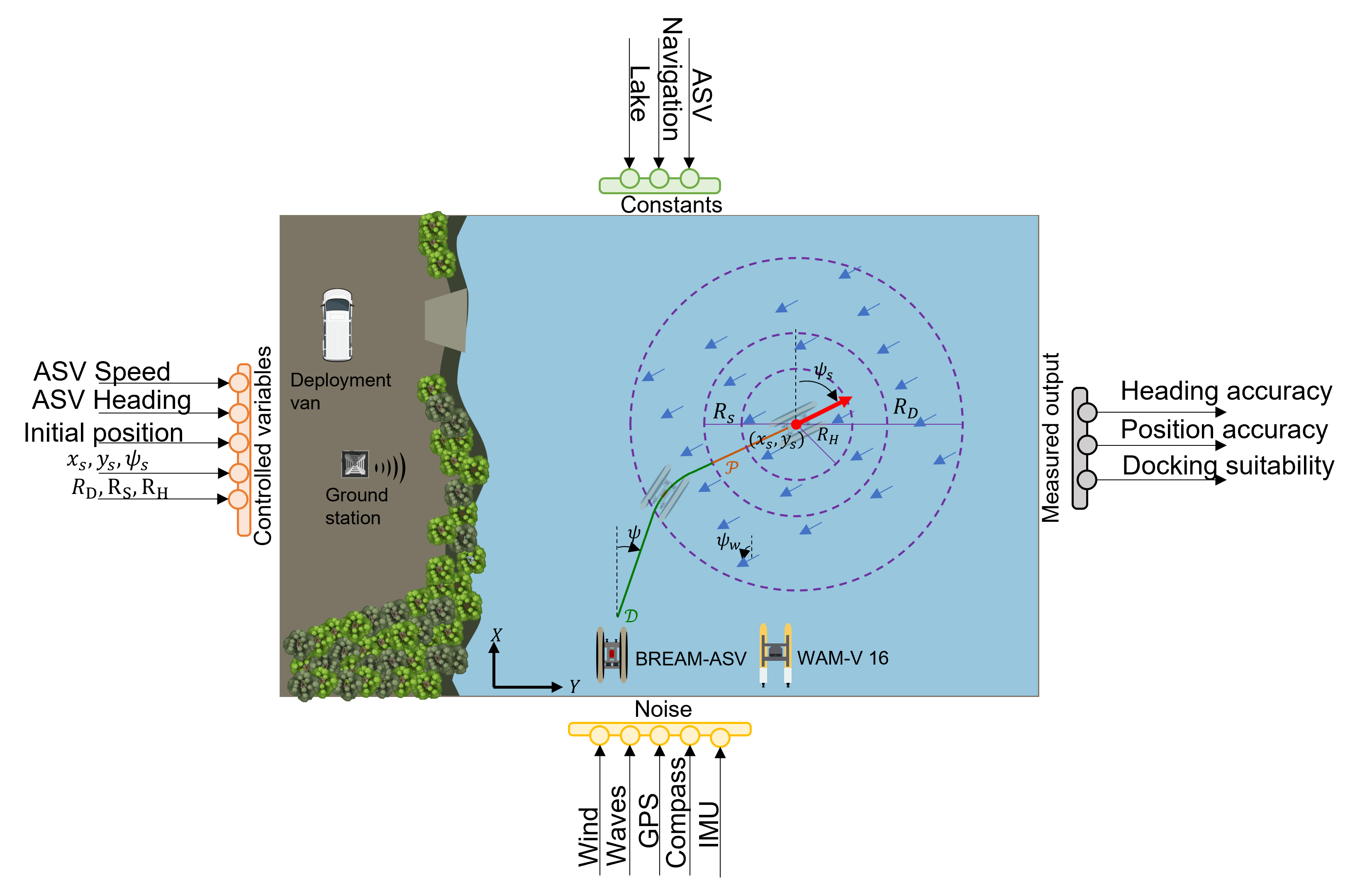}
\caption{This diagram shows the experimental setup. It includes the main components used during the deployment and the testing. Additionally, it shows the controlled variables during the experiment, the constant elements, as well as the noise present during the test. The measured output block shows the elements used to validate the accuracy of the overall approach.}
\label{fig:test_setup}
\end{figure}

To evaluate the station keeping strategy, several tests were defined to observe the performance under different wind directions. Additionally, we performed a test to verify the performance when the wind sensor is not present. For the first validation stage, a total of 60 hours were spent on Fairfield Lakes, IN for deployment, debugging, data collection, and fine-tuning. 5 tests under the wind conditions are reported in Table \ref{tbl:mpc_perf_real_new}. For all the tests the initial position of the ASV was larger than $R_D=20m$ at a random heading. For the second stage of validation, a total of 32 hours were spent on Lake Harner, IN. 5 tests under the wind conditions are reported in Table \ref{tbl:mpc_perf_real_wamv}. For all the tests the vehicle was started at a random initial position larger than $R_D=20m$ with a random initial heading. Fig.~\ref{fig:test_setup} shows the experimental test setup employed for the experiments run on BREAM-ASV and WAM-V 16. 

\begin{figure}[t]
\vspace{0.3cm}
\centering
\includegraphics[width=0.5\columnwidth]{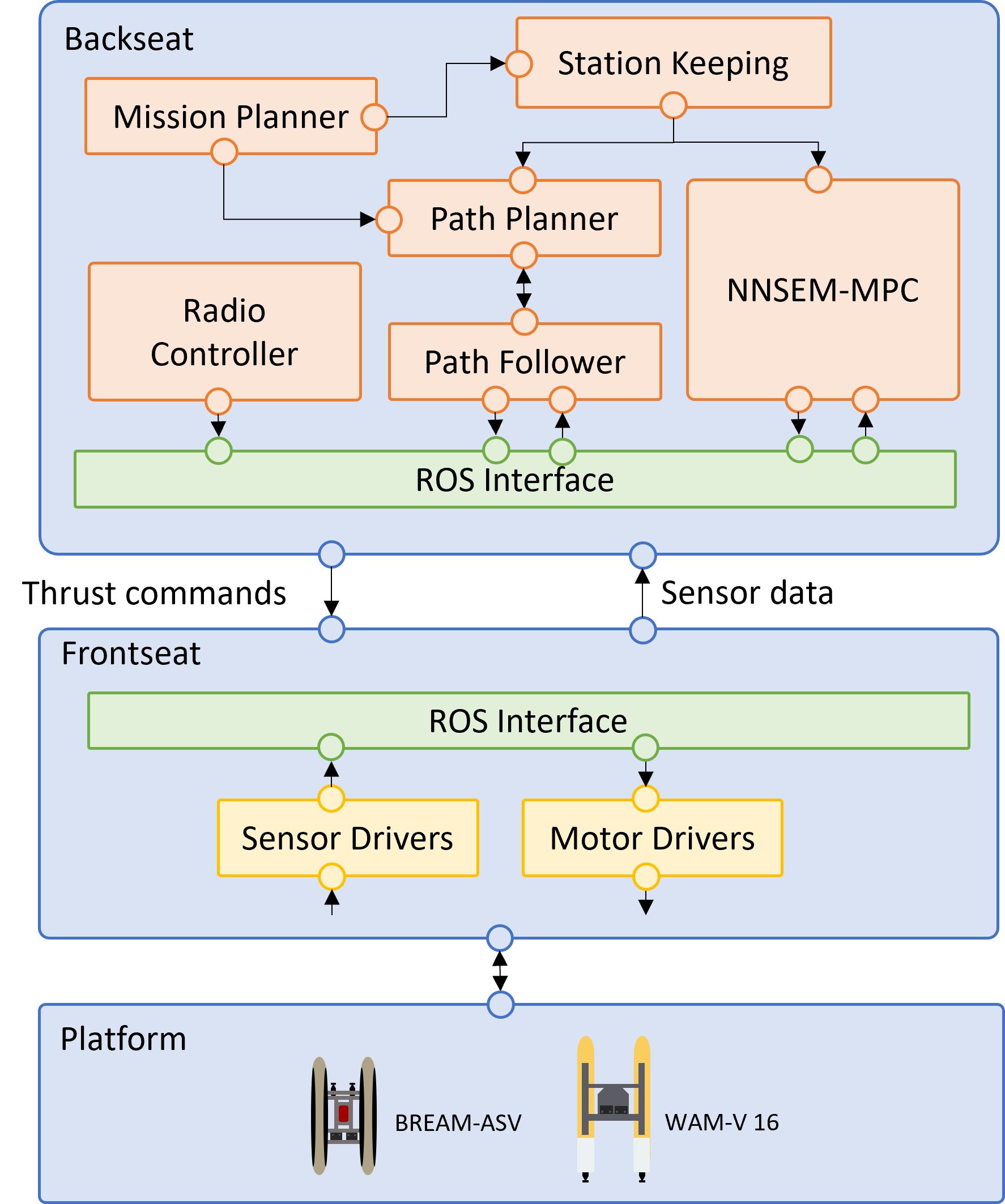}
\caption{This system diagram illustrates the usage of the backseat frontseat architecture for the purpose of the implementation of the NNSEM-MPC approach. Backseat implements the behavior-based controller, while the frontseat is in charge of the low-level controllers.}
\label{fig:system_diagram}
\end{figure}

A backseat-frontseat architecture was adopted to enhance the modularity and portability of our NNSEM-MPC controller, this is a common choice in marine vehicles. In this configuration, behavior-based controllers reside on the backseat side, while real-time low-level controllers operate on the frontseat side. These backseat and frontseat components communicate through ROS, facilitating straightforward implementation, even when multiple computers are involved. This architectural design allows for the easy deployment of the NNSEM-MPC approach across various platforms, irrespective of their underlying differences. We illustrate this adaptability in our work using the BREAM-ASV and WAM-V 16 platforms. Fig.~\ref{fig:system_diagram} shows the layered architecture of the system and the components involved in implementing the NNSEM-MPC controller.

\subsection{Experimental Validation I: Proof of Concept on BREAM-ASV}


The first stage of the experimental validation was carried out on the BREAM-ASV, which is a differential drive ASV featuring a backseat-frontseat architecture: one Raspberry Pi and one Jetson Nano, where the Jetson Nano is the backseat in charge of the overall mission planning, the MPC optimization, and the neural network model. The Raspberry Pi is the frontseat which implements the low-level drivers. All of the computers are connected over Ethernet and run under the ROS framework. The purpose of this initial validation was to integrate all the components of the NNSEM-MPC into a real target and figure out the challenges or improvements required for a subsequent iteration.


\begin{figure}
     \centering
     \begin{subfigure}[b]{0.35\textwidth}
         \centering
         \includegraphics[width=\textwidth]{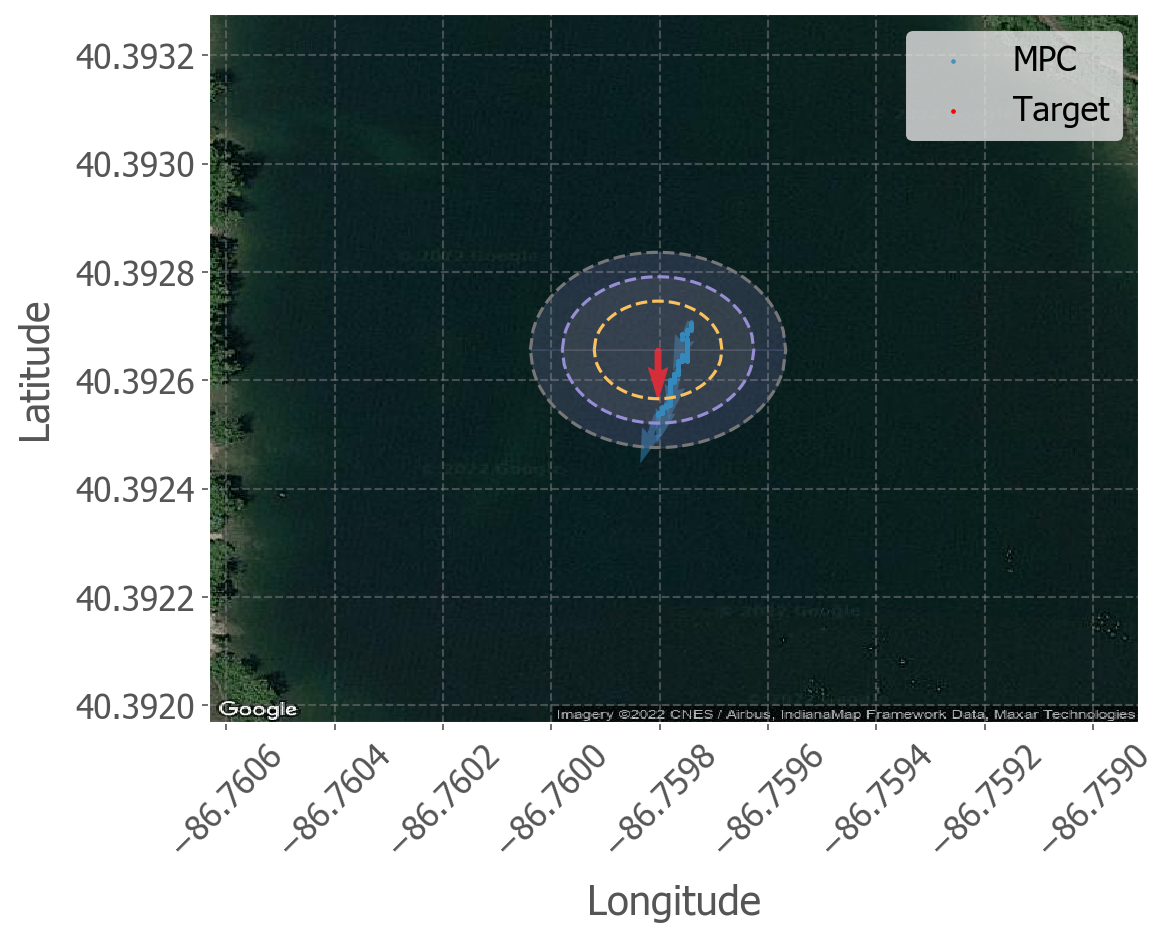}
         \caption{$ $}
         \label{fig:sk_test_window_1_1}
     \end{subfigure}
     \hspace{5pt}
     \begin{subfigure}[b]{0.35\textwidth}
         \centering
         \includegraphics[width=\textwidth]{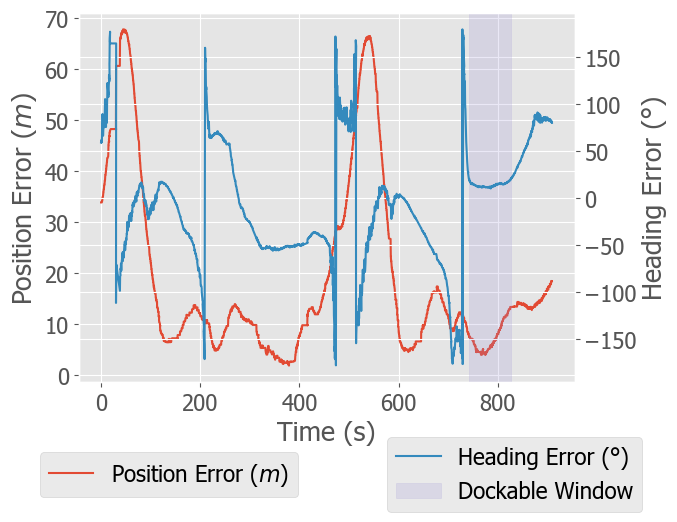}
         \caption{$ $}
         \label{fig:dockable_window_1}
     \end{subfigure}
     \hfill
     \begin{subfigure}[b]{0.35\textwidth}
         \centering
         \includegraphics[width=\textwidth]{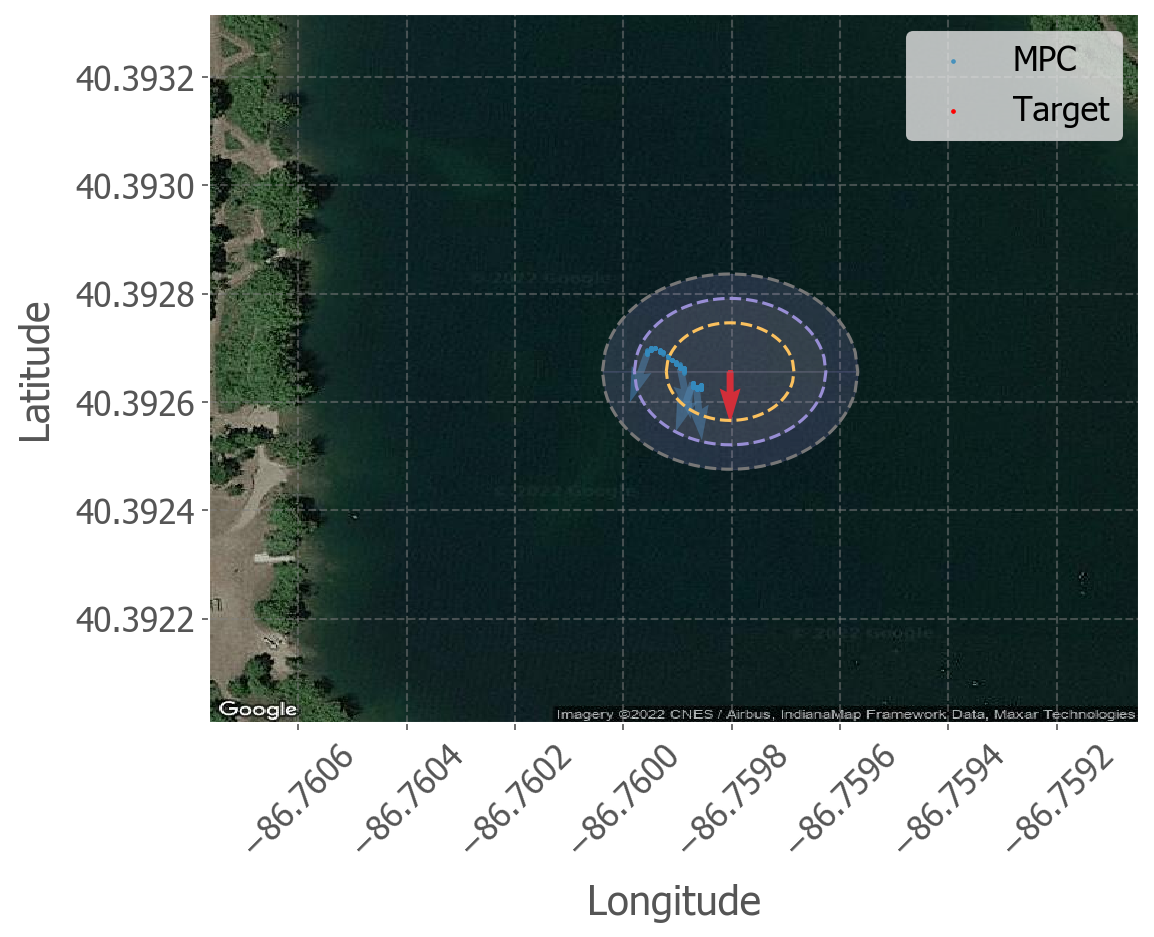}
         \caption{$ $}
         \label{fig:sk_test_window_7_1}
     \end{subfigure}
     \hspace{5pt}
     \begin{subfigure}[b]{0.35\textwidth}
         \centering
         \includegraphics[width=\textwidth]{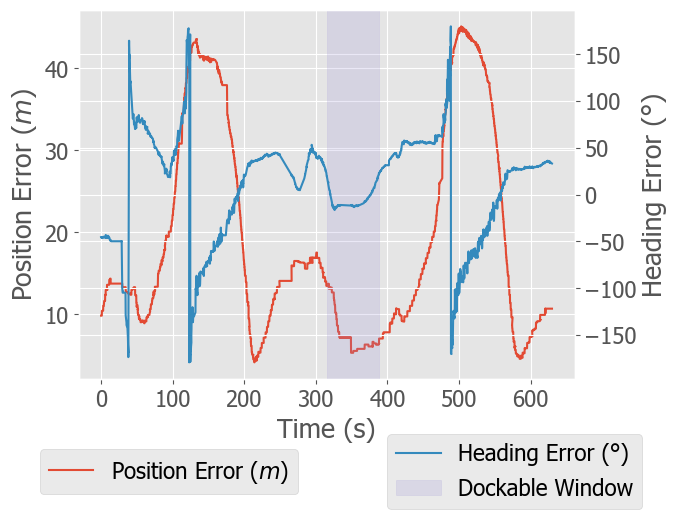}
         \caption{$ $}
         \label{fig:dockable_window_7}
     \end{subfigure}
 
    \caption{This figure shows the trajectories from all the tests carried out suitable used for underwater docking, where a) and b) correspond to a segment of 132.4 s from test 1, and c) and d) correspond to 56.6 s from test 4. In these tests $R_{D}=20,R_{s}=15m,$ and $R_{H}=10$.}
    \label{fig:dockable_windows}
\end{figure}

To evaluate the station keeping strategy several tests were defined to observe the performance under different wind directions. Additionally, a couple of tests were defined to verify the performance when no MPC controller is active, and also when the wind sensor is not present. In total, five tests were carried out on Fairfield Lakes, IN the same day, under the wind conditions reported in Table \ref{tbl:mpc_perf_real_new}. For these field experiments, the time horizon $T$ was reduced from $4$ to $1.6$ in order to fit the MPC processing time within the control loop period which for BREAM-ASV is $0.2$s.

A summary is reported in Table \ref{tbl:mpc_perf_real_new}, where tests 1 through 3 were run using the same MPC optimization weights and the same time horizon of $T=1.6$ s. Tests 4 and 5 used a different set of optimization weights. For test 5 the wind sensor was disabled. Overall, the performance of the station keeping strategy can be evaluated based on the distance error and the heading error. The position error for all the tests is always below $13$ m, and the heading error varies according to the target heading. Overall, the best results are obtained when the ASV keeps its position against the wind, which can be observed in the lower heading error and standard deviation. The standard deviation of the heading error is an aspect to look at since a more steady heading will increase the chances for underwater docking.

Following the latter tests 1 and 4 not only have a lower heading error, but they have a lower standard deviation of the heading error, which implies a more steady convergence to the desired configuration $\pmb{\eta}_s$. When the ASV is perpendicular to the wind, we notice an increase in both the position and heading error. This is an expected behavior since the ASV does not have a way to counteract the effect of lateral disturbance besides re-planning as described in Sec.~\ref{ssec:sk_nn_model}.
\renewcommand{\tabcolsep}{1.5pt}
\begin{table}[]
\centering
\begin{tabular}{lllllllllll}
\cline{2-11}
                & \multirow{3}{*}{\textbf{\begin{tabular}[c]{@{}l@{}}Wind\\ Sensor\end{tabular}}} & \multirow{3}{*}{\textbf{\begin{tabular}[c]{@{}l@{}}Target\\ Head(°)\end{tabular}}} & \multicolumn{2}{l}{\multirow{2}{*}{\textbf{\begin{tabular}[c]{@{}l@{}}Distance \\ Error (m)\end{tabular}}}} & \multicolumn{2}{l}{\multirow{2}{*}{\textbf{\begin{tabular}[c]{@{}l@{}}Heading \\ Error (°)\end{tabular}}}} & \multicolumn{2}{l}{\multirow{2}{*}{\textbf{\begin{tabular}[c]{@{}l@{}}Wind \\ Speed (m/s)\end{tabular}}}} & \multicolumn{2}{l}{\multirow{2}{*}{\textbf{\begin{tabular}[c]{@{}l@{}}Wind \\ Heading (°)\end{tabular}}}} \\
                &                                                                                 &                                                                                    & \multicolumn{2}{l}{}                                                                                        & \multicolumn{2}{l}{}                                                                                       & \multicolumn{2}{l}{}                                                                                      & \multicolumn{2}{l}{}                                                                                      \\ \cline{4-11} 
                &                                                                                 &                                                                                    & \textbf{Mean}                                         & \textbf{Std}                                        & \textbf{Mean}                                        & \textbf{Std}                                        & \textbf{Mean}                                        & \textbf{Std}                                       & \textbf{Mean}                                        & \textbf{Std}                                       \\ \hline
\textbf{Test 1} & Yes                                                                             & 180                                                                                & 9.86                                                  & 4.25                                                & -8.71                                                & 57.26                                              & 2.20                                                 & 0.78                                               & 15.02                                                & 36.69                                              \\
\textbf{Test 2} & Yes                                                                             & 0                                                                                  & 11.24                                                 & 4.01                                                & 23.97                                                & 80.67                                              & 2.41                                                 & 0.79                                               & 7.82                                                & 35.61                                              \\
\textbf{Test 3} & Yes                                                                             & 90                                                                                 & 10.14                                                 & 4.97                                                & 44.53                                                & 71.73                                               & 2.31                                                 & 1.12                                               & 10.90                                                 & 31.19                                              \\
\textbf{Test 4} & Yes                                                                             & 180                                                                                & 11.34                                                 & 4.00                                                & 28.69                                                & 34.69                                               & 2.24                                                 & 0.69                                               & -19.31                                                & 29.70                                              \\
\textbf{Test 5} & No                                                                              & 180                                                                                & 12.01                                                 & 3.89                                                & 13.25                                                & 64.35                                               & 2.27                                                 & 0.78                                               & -16.43                                                & 24.71                                              \\ \hline
\end{tabular}

\caption{Performance of the MPC controller for each of the tests and wind conditions during the deployment on Fairfield Lakes, IN.}
\label{tbl:mpc_perf_real_new}
\vspace{-0.6cm}
\end{table}

In order to assess the suitability of the method to improve underwater docking, it is important to define the metrics that allow underwater docking. In general underwater docking is done on stationary platforms or to a slowly moving platform \cite{yazdani2020survey,page2021underwater}; therefore, it is desirable to keep ASV's linear and rotational speeds bounded (0.5 m/s and 5 $^{\circ}$/s). In addition to that, we consider a circular sector of 15 m of radius and 40$^{\circ}$ centered at the target heading. Fig. \ref{fig:dockable_windows} shows two time windows in which the ASV reaches a steady state that could be used to enable underwater docking to a mobile target. The mean position error of tests 1 and 4 in these docking windows is 7.87 m and 8.05 m, while the heading error is 13.97 $^{\circ}$ and 8.05 $^{\circ}$, respectively.

Overall, the results obtained in this first stage validated the concept, showcasing its potential. However, we encountered that the GPS and compass sensors used on BREAM-ASV were not only noisy but had a low precision and inconsistent output. This was reflected in abrupt heading or position readings, which made the NNSEM-MPC approach struggle to find an optimal solution. On the other hand, the MPC solver \verb|mpc.torch| although easy to integrate, exhibited a slow performance which made every iteration of the MPC lag resulting in delayed control actions and poor station keeping performance in terms of position and heading errors. 

As we prepared for the second stage, while the BREAM-ASV was capable of running all the required software components from a computational standpoint, we made the decision to transition to the WAM-V 16 platform. This choice was primarily driven by the WAM-V's reliable sensor package, which we believe will substantially improve the accuracy and overall performance of the system.


\subsection{Experimental Validation II: Refined Implementation on WAM-V16}

The second stage of the experimental validation was carried out on the WAM-V 16 platform which is a commercial surface vehicle with independent articulated pontoons designed to keep a steady deck motion regardless of the environmental conditions. It features a VN-300 Dual GNSS/INS capable of providing stable position, velocity, and acceleration of the system. WAM-V 16 has the same backseat-frontseat architecture, where the frontseat runs the lower level controller and the backseat runs the mission planner and the MPC. The WAM-V 16 as a commercial off-the-shelf vehicle has pre-built maneuvers including station keeping. However, it does not take the heading into consideration and only allows the vehicle to maintain its position with a large tolerance. As a first step, we validated the sensor package available on WAM-V 16, verifying the correctness of the GPS and compass values, as well as the associated noise. Additionally, a custom CasADi based module was developed to optimize the NN model within the MPC optimization loop, which helped to improve the speed of the system, thus, making the real-time implementation possible.

\begin{figure}[!ht]
\centering
\vspace{0.2cm}
\includegraphics[width=0.9\columnwidth]{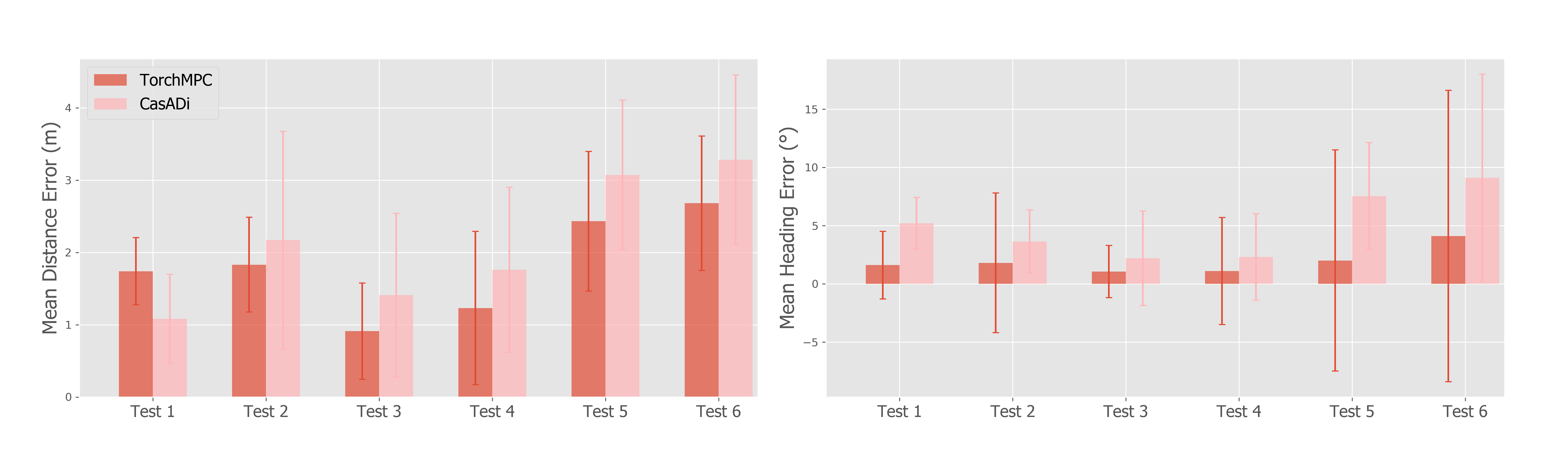}
\caption{Comparison of the MPC solvers using the same NNSEM model for 6 tests in simulation. The test settings are the same as the settings in section \ref{ssec:nn_eval}. The peaks of the bars represent the mean and the ends of the error bars depict $\pm 0.5$ standard error.}
\label{fig:solver_compare}
\vspace{-0.3cm}
\end{figure}

\begin{figure}
     \centering
     \begin{subfigure}[b]{0.35\textwidth}
         \centering
         \includegraphics[width=\textwidth]{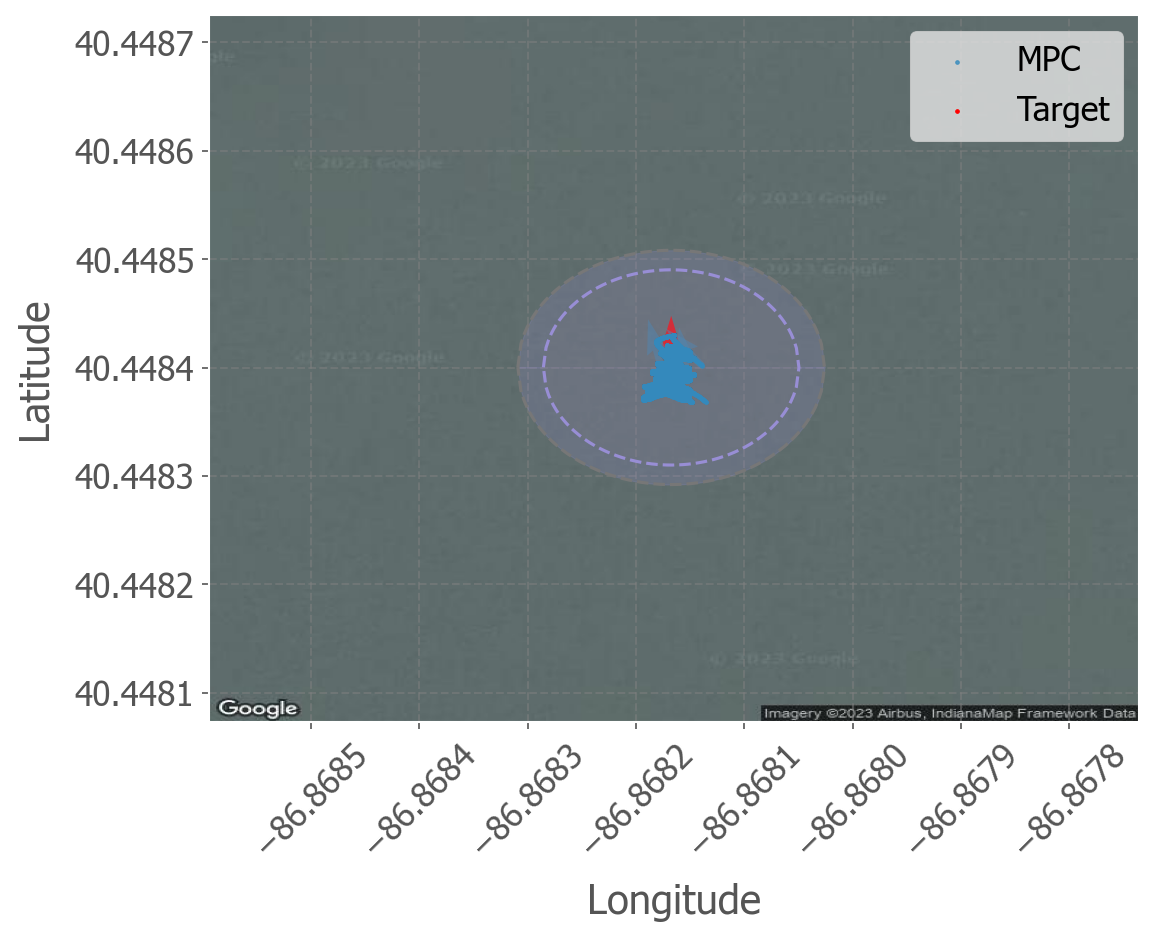}
         \caption{$ $}
         \label{fig:wamv_sk_test_window_north_1}
     \end{subfigure}
     \hspace{5pt}
     \begin{subfigure}[b]{0.35\textwidth}
         \centering
         \includegraphics[width=\textwidth]{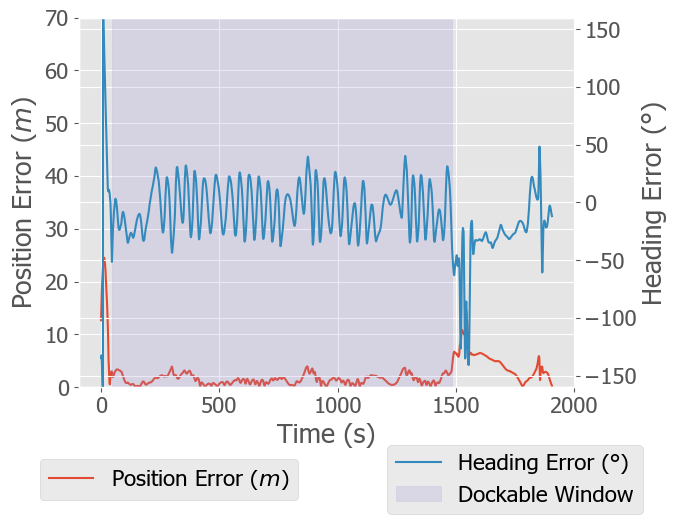}
         \caption{$ $}
         \label{fig:wamv_dockable_window_north_1}
     \end{subfigure}
     \hfill
     \begin{subfigure}[b]{0.35\textwidth}
         \centering
         \includegraphics[width=\textwidth]{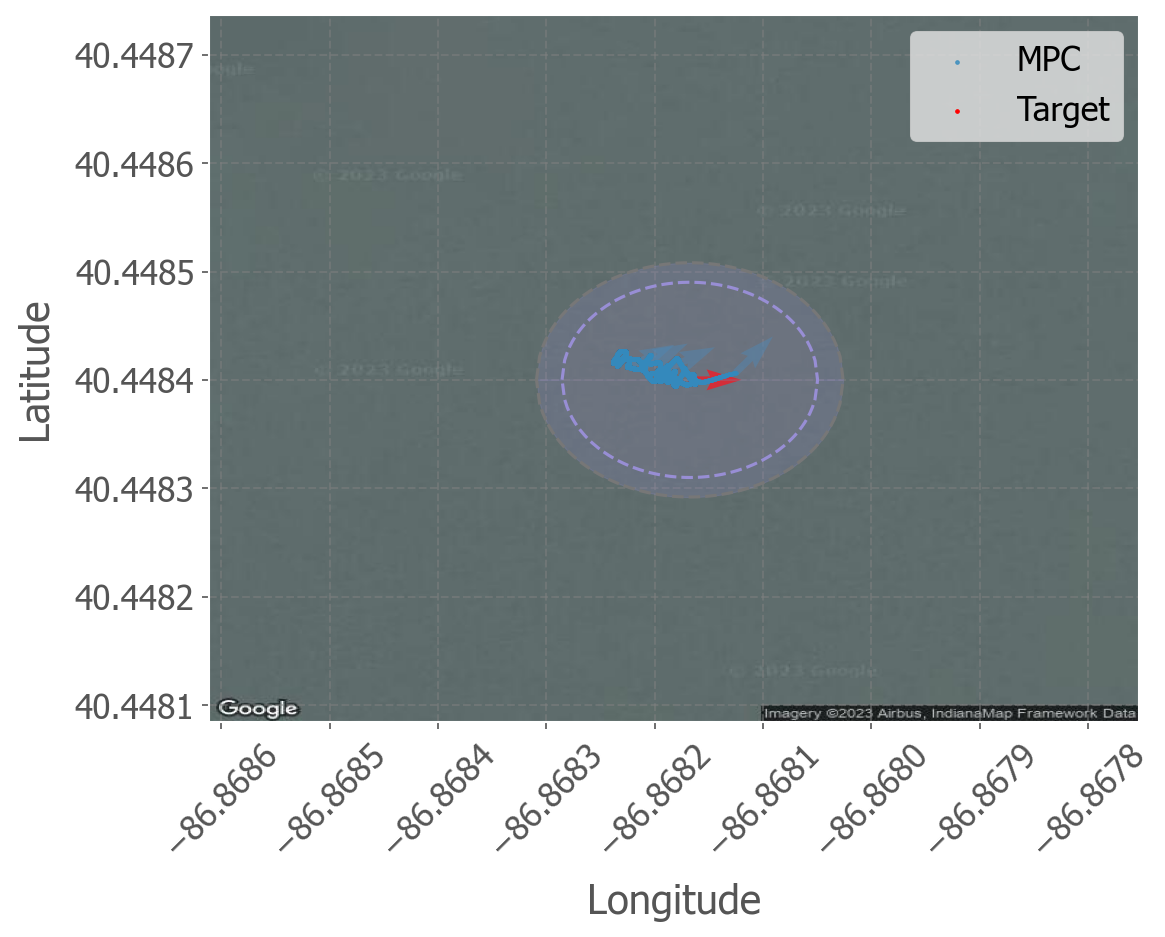}
         \caption{$ $}
         \label{fig:wamv_sk_test_window_east_1}
     \end{subfigure}
     \hspace{5pt}
     \begin{subfigure}[b]{0.35\textwidth}
         \centering
         \includegraphics[width=\textwidth]{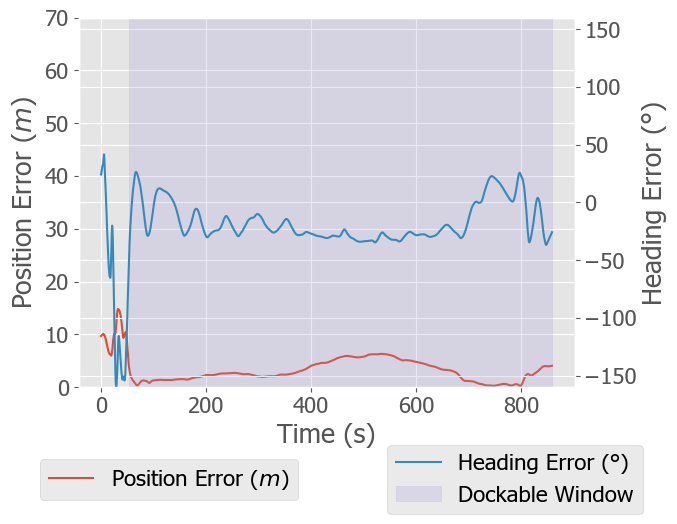}
         \caption{$ $}
         \label{fig:wamv_dockable_window_east_3}
     \end{subfigure}
     \begin{subfigure}[b]{0.35\textwidth}
         \centering
         \includegraphics[width=\textwidth]{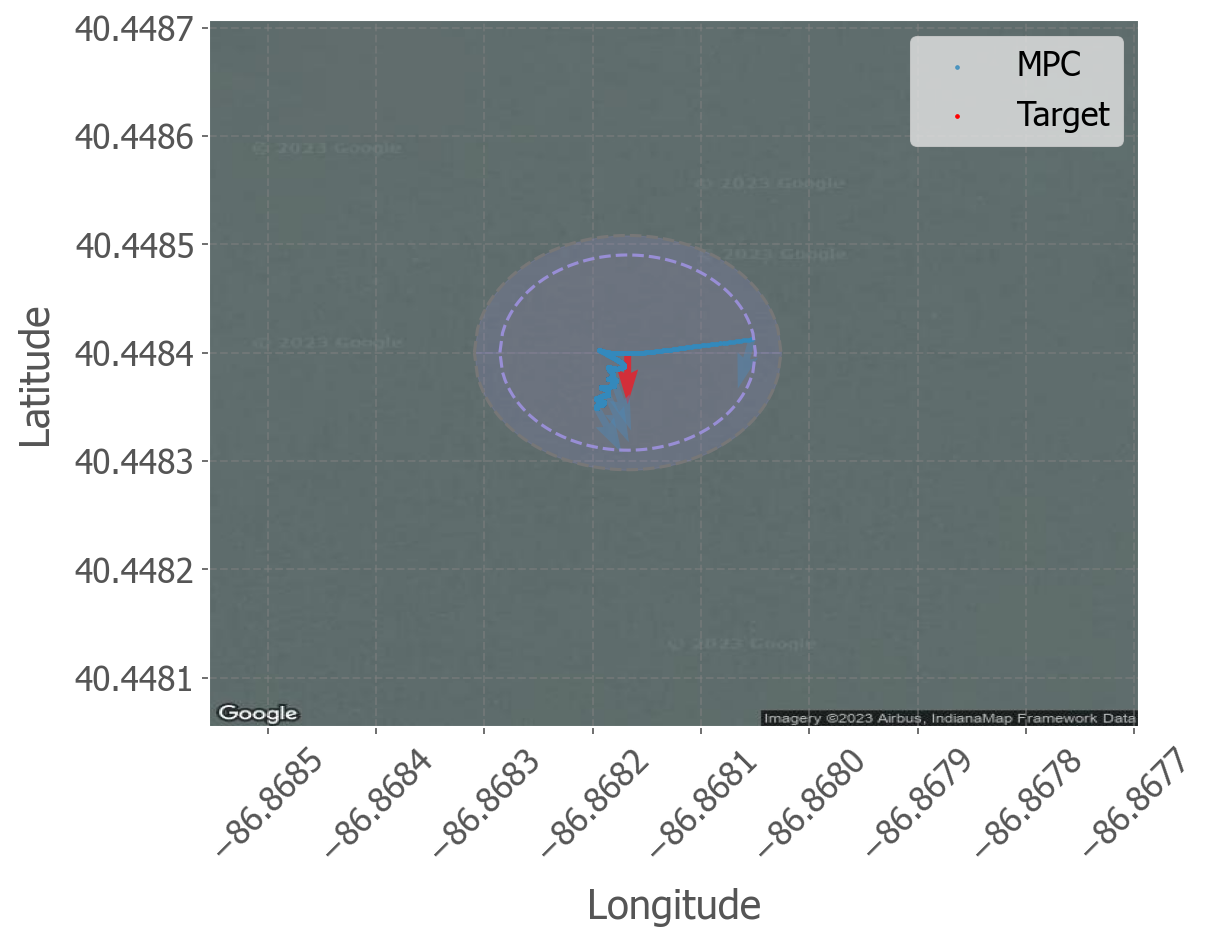}
         \caption{$ $}
         \label{fig:wamv_sk_test_window_south_1}
     \end{subfigure}
     \hspace{5pt}
     \begin{subfigure}[b]{0.35\textwidth}
         \centering
         \includegraphics[width=\textwidth]{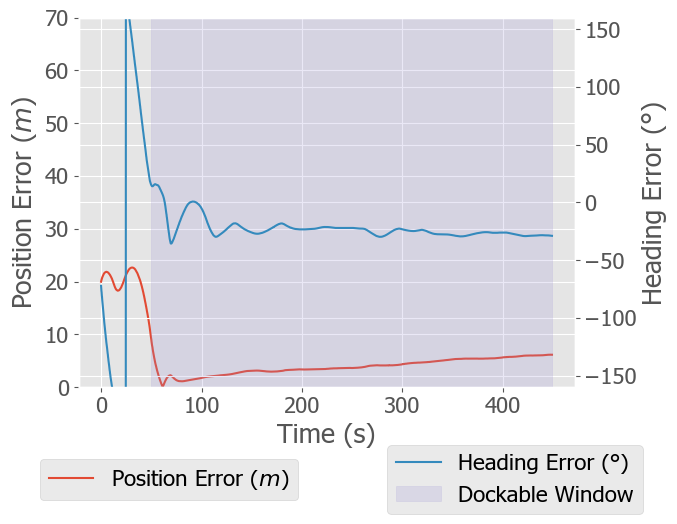}
         \caption{$ $}
         \label{fig:wamv_dockable_window_south_1}
     \end{subfigure}
     \hfill
     \begin{subfigure}[b]{0.35\textwidth}
         \centering
         \includegraphics[width=\textwidth]{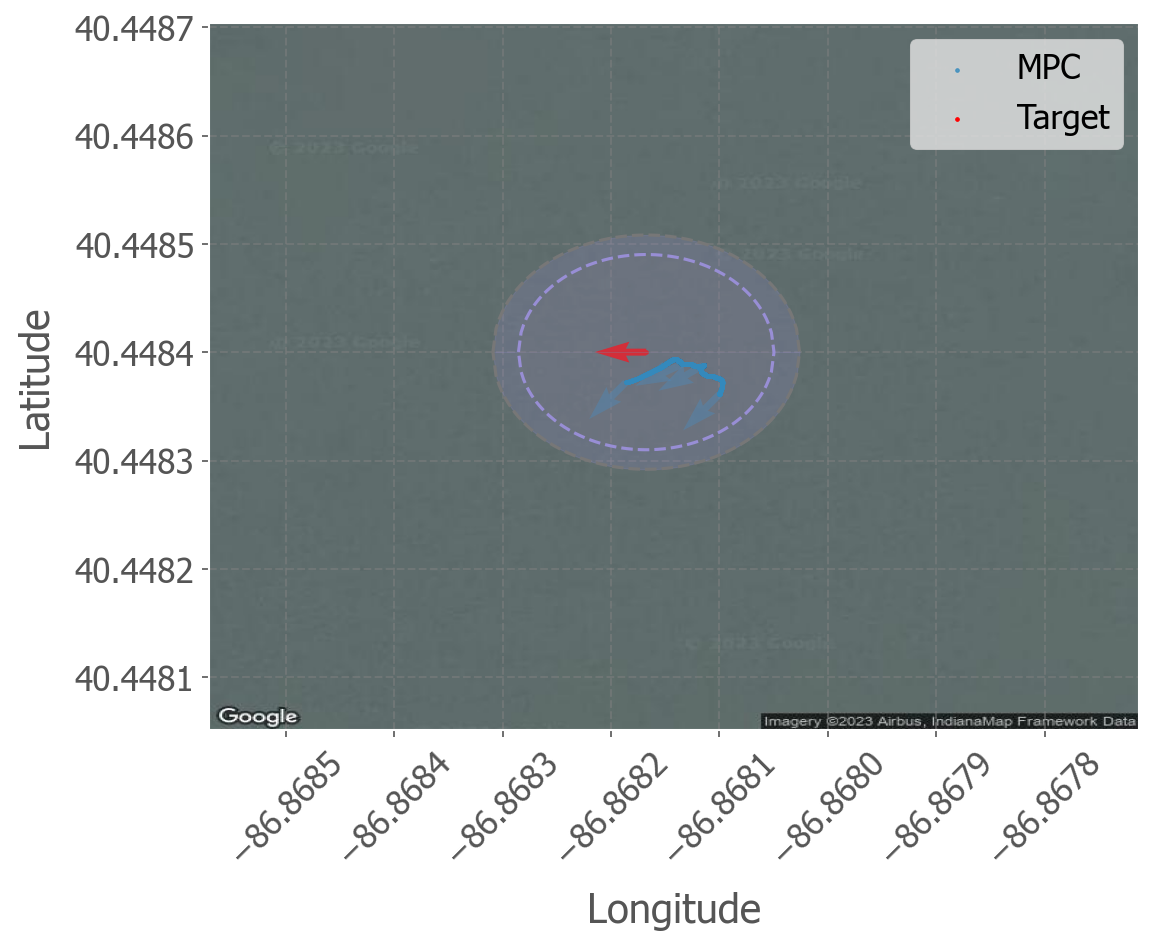}
         \caption{$ $}
         \label{fig:wamv_sk_test_window_west_1}
     \end{subfigure}
     \hspace{5pt}
     \begin{subfigure}[b]{0.35\textwidth}
         \centering
         \includegraphics[width=\textwidth]{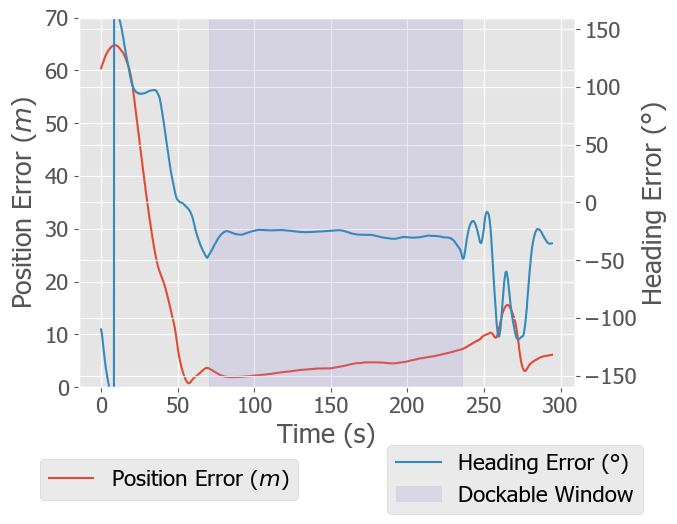}
         \caption{$ $}
         \label{fig:wamv_dockable_window_west_1}
     \end{subfigure}
        \caption{This figure shows the trajectories from all the tests carried out in Lake Harner, that are suitable for underwater docking. In general, all the test cases exhibited long docking windows of time $>200$s in the presence of wind disturbances. In these tests $R_{D}=20,R_{s}=10m,$ and $R_{H}$ is not used.}
        \label{fig:dockable_windows_wamv}
\end{figure}
\renewcommand{\tabcolsep}{1.5pt}
\begin{table}[]
\centering
\begin{tabular}{lllllllllll}
\cline{2-11}
                & \multirow{3}{*}{\textbf{\begin{tabular}[c]{@{}l@{}}Wind\\ Sensor\end{tabular}}} & \multirow{3}{*}{\textbf{\begin{tabular}[c]{@{}l@{}}Target\\ Head(°)\end{tabular}}} & \multicolumn{2}{l}{\multirow{2}{*}{\textbf{\begin{tabular}[c]{@{}l@{}}Distance \\ Error (m)\end{tabular}}}} & \multicolumn{2}{l}{\multirow{2}{*}{\textbf{\begin{tabular}[c]{@{}l@{}}Heading \\ Error (°)\end{tabular}}}} & \multicolumn{2}{l}{\multirow{2}{*}{\textbf{\begin{tabular}[c]{@{}l@{}}Wind \\ Speed (m/s)\end{tabular}}}} & \multicolumn{2}{l}{\multirow{2}{*}{\textbf{\begin{tabular}[c]{@{}l@{}}Wind \\ Heading (°)\end{tabular}}}} \\
                &                                                                                 &                                                                                    & \multicolumn{2}{l}{}                                                                                        & \multicolumn{2}{l}{}                                                                                       & \multicolumn{2}{l}{}                                                                                      & \multicolumn{2}{l}{}                                                                                      \\ \cline{4-11} 
                &                                                                                 &                                                                                    & \textbf{Mean}                                         & \textbf{Std}                                        & \textbf{Mean}                                        & \textbf{Std}                                        & \textbf{Mean}                                        & \textbf{Std}                                       & \textbf{Mean}                                        & \textbf{Std}                                       \\ \hline
\textbf{Test 1} & Yes    & 0     & 2.08   & 1.81   & -9.19   & 23.05   & 1.64   & 1.01   & 98.11     & 44.63   \\
\textbf{Test 2} & Yes    & 90    & 3.21   & 2.16   & -18.24  & 21.54   & 1.14   & 0.82   & 94.21     & 48.16   \\
\textbf{Test 3} & Yes    & 180   & 3.89   & 1.46   & -22.48  & 8.99    & 0.88   & 1.07   & 95.38     & 36.62   \\
\textbf{Test 4} & Yes    & 270   & 4.52   & 2.17   & -30.18  & 18.09   & 2.11   & 1.94   & 127.61    & 63.79   \\ \hline
\end{tabular}

\caption{Performance of the MPC controller for each of the tests and wind conditions during the deployment on Lake Harner, IN.}
\label{tbl:mpc_perf_real_wamv}
\vspace{-0.6cm}
\end{table}

To validate the performance of the CasADi based MPC solver we compare it against the \verb|mpc.torch| based MPC solver, six tests have been conducted in simulation. The test settings are consistent with those outlined in Section \ref{ssec:nn_eval}. Tests $1$, $3$, and $5$ involve mild wind disturbances, while tests $2$, $4$, and $6$ incorporate stronger disturbances. For tests $1$ and $2$, the goal heading aligns with the wind direction; for tests $3$ and $4$, the goal heading opposes the wind; and for tests $5$ and $6$, the goal heading is perpendicular to the wind. The testing results are shown in Fig. \ref{fig:solver_compare} and indicate that the CasADi solver exhibits larger distance and heading errors in most of the tests. However, the mean iteration times of the first $10$ steps for NNSEM-MPC using \verb|mpc.torch| is $0.68$ s while CasADi is only $0.18$ s. This time performance difference is mainly because CasADi uses a variety of numerical optimal control methods quickly and effectively. The observed differences in control performance can be attributed to multiple factors. One of them is the better linearization done by \verb|mpc.torch|. Another factor is the the precision loss while loading NN parameters onto CasADi. Finally the larger number of parameters to tune for the CasADi MPC solver. Overall, the differences are statistically not significant, and therefore the improved computational performance of CasADi was used for the implementation of the MPC solver on the WAM-V 16 platform.


For the second stage test, the wind conditions were generally milder than those in the first stage test. However, the wind direction exhibited more variability throughout the trials, as indicated by the larger wind direction standard deviation. The second stage comprised four tests, where we aimed to align the vehicle both in the direction of the wind and perpendicular to it, allowing for a more comprehensive assessment of performance. In this manner, tests 1 (north) and 3 (south) were conducted with the wind blowing perpendicular to the target heading, while tests 2 (east) and 4 (west) were designed to have the wind aligned with and against the target heading, respectively.

Overall, tests 1 and 3 in Table~\ref{tbl:mpc_perf_real_wamv} exhibited the best performance when considering the penalty score ($ps$). However, this improvement came at the cost of a more energy-intensive behavior, as shown in Fig.\ref{fig:dockable_windows_wamv}. This outcome supports the intuition described in \cite{pereira2008experimental} that maintaining position in the presence of disturbances acting on an uncontrolled axis (the lateral axis) requires the vehicle to maneuver in a way that counters position errors but compromises heading stability. For the specific case of test 3 this latter, might not apply completely, however, the wind speed during this test was the lowest of all the test scenarios, which may explain the good performance. On the other hand, tests 2 and 4 from Table~\ref{tbl:mpc_perf_real_wamv}, resulted in a higher penalty score but demonstrated a more stable behavior as depicted in Fig.\ref{fig:dockable_windows_wamv}. This finding reinforces the results obtained in \cite{pereira2008experimental} that controlling the vehicle along or against its direction not only promotes stability but also enhances energy efficiency. This is a desired behavior to improve a higher docking success rate.

The results in this Section demonstrated that the second stage tests on WAM-V 16 ASV resulted not only in lower position and heading errors, but also in longer and more stable docking windows as depicted in Fig.~\ref{fig:dockable_windows_wamv} when compared to the first stage tests on BREAM-ASV. The sensor package plays an essential role in the accuracy of the method, first to improve the system identification, and second to better estimate the optimal action within the MPC controller. The efficiency of the MPC solver allows the necessary reactivity to environmental changes.
The field experiments showed that the ASV can effectively keep the station utilizing the proposed method, with a position error as low as 1.68 m and a heading error as low as $6.14^{\circ}$ within the docking time windows of at least 150 s.

\section{CONCLUSIONS}\label{sec:conc} 

In this paper, NNSEM-MPC, a framework incorporating a neural network simulation error minimization model and MPC is proposed and applied in ASV station keeping scenarios. To validate the performance of the proposed approach, nonlinear controllers such as backstepping controller and sliding mode controller, as well as data-driven controllers such as NODE-MPC, and KNODE-MPC are used as baselines to compare through simulation and experimental validation in Gazebo. 

Additionally, NNSEM-MPC is deployed on the two ASV platforms to evaluate its performance under real conditions. In order to account for the disturbances, an anemometer is integrated. Once the sensor is integrated, data collection is carried out in simulation and field deployments to train a neural network model that accurately predicts the behavior of the ASVs. Filtering and sensor fusion are required to correctly predict the surge and sway of the ASV, which impacts the MPC performance. The results obtained from the simulation clearly show the advantage of MPC over the other controllers, especially due to the accuracy of the trained neural network model to predict the actual vehicle's state. In all the simulation test cases the proposed approach reaches a position error of less than $2$ m and a heading error of less than $2^\circ$ when the target heading is opposed or along the wind direction. In contrast, the error increases up to $3$ m and $5^\circ$ when the wind is perpendicular to the target heading. 

For the experimental validation was conducted in two stages on BREAM-ASV and WAM-V 16. The experiments in the first stage using BREAM-ASV validated the idea of the proposed method, but the performance was rather poor, mainly due to the noisy sensors and slow MPC solver. The first stage field experiments provided insight  the potential areas of improvement for the implementation. In the second stage of validation with the WAM-V 16 platform, we made significant enhancements. This included improving the sensor quality and optimizing the MPC solver, leading to a better system model and a faster control loop. These improvements  resulted in a mean position error consistently below $4.52$ meters across all test cases. Within docking windows WAM-16 reached a position and heading errors as low as $1.68$ m and $6.14^\circ$, respectively. In 4 tests, the docking windows were all larger than $150$ s.



Further development plans include improving the system model considering wave disturbances in the state and collecting data in various environmental and vehicle conditions. This would improve the ability of the NNSEM-MPC to better estimate future states and generate optimal control commands. Future applications of this project will be extensive. The ASV capable of keeping a station has the potential to collaborate with other robots from different domains including UAVs and AUVs. The approach will be used as part of the autonomous replenishment process, for that, an ASV will be equipped with a docking system stationary for aerial and undersea vehicles to get docked and recharged. This will enable the long-term operation of heterogeneous autonomous systems with minimal human intervention.




\bibliographystyle{IEEEtran}
\bibliography{station_keeping.bib}
 
\end{document}